%% file: main.tex
\documentclass{article} % For LaTeX2e
\usepackage{iclr2026_conference,times}

\input{math_commands.tex}

\input{preamble}

\definecolor{cvprblue}{rgb}{0.21,0.49,0.74}

\usepackage{hyperref}
\usepackage{url}

\title{
Learning Patient-specific Disease Dynamics with Latent Flow Matching for Longitudinal Imaging Generation
}

\author{
Hao Chen$^{1,}$\thanks{Equal contribution. Contact: hc666@cam.ac.uk.} \quad
Rui Yin$^{2,*}$ \quad
Yifan Chen$^{1}$ \quad
Qi Chen$^{3}$ \quad
Chao Li$^{1,4,}$\thanks{Corresponding author.} \\
$^{1}$University of Cambridge, \quad
$^{2}$Nanjing First Hospital, Nanjing Medical University \\
$^{3}$Johns Hopkins University \quad 
$^{4}$University of Dundee \\
}

\iclrfinalcopy % Uncomment for camera-ready version, but NOT for submission.

\begin{document}

\maketitle

\input{sec/0_abstract}

\input{sec/1_intro}

\input{sec/2_related}

\input{sec/3_method}
\input{sec/4_experiment}

\input{sec/5_conclusion}

{
    \small
    \bibliography{main}
    \bibliographystyle{iclr2026_conference}
}

\input{sec/6_appendix}

\end{document}

%% file: math_commands.tex
\usepackage{amsmath,amsfonts,bm}

\def\eqref#1{equation~\ref{#1}}
\def\1{\bm{1}}

\DeclareMathAlphabet{\mathsfit}{\encodingdefault}{\sfdefault}{m}{sl}
\SetMathAlphabet{\mathsfit}{bold}{\encodingdefault}{\sfdefault}{bx}{n}

%% file: preamble.tex
\usepackage{amsmath}
\usepackage{mathtools}    % for   \coloneqq

\usepackage{multirow} 
\usepackage{booktabs}
\usepackage{enumitem}

\usepackage[table]{xcolor}

\usepackage{graphicx}
\usepackage{amsmath}
\usepackage{amssymb}
\usepackage{booktabs}
\usepackage{pifont}

\usepackage{algorithmicx}
\usepackage{algorithm}
\usepackage{algpseudocode}

\usepackage{graphicx}
\usepackage{caption}
\usepackage{subcaption}  % instead of \caption*

\colorlet{framecolor}{black!30}       % dark gray instead of pure black

\usepackage[dvipsnames]{xcolor}
\definecolor{cvprblue}{rgb}{0.21,0.49,0.74}
\definecolor{Mahogany}{RGB}{192,64,0}
\definecolor{CadetBlue}{rgb}{0.3725, 0.6196, 0.6275}

\usepackage[pagebackref]{hyperref}
\hypersetup{
	final,
	breaklinks,
	colorlinks,
	linkcolor={Mahogany},
	citecolor={cvprblue},
	urlcolor={CadetBlue}
}

\usepackage{cleveref}

\newcommand{\eg}[0]{\textit{e.g.},}
\newcommand{\ie}[0]{\textit{i.e.},}
\newcommand{\etal}[0]{\textit{et al.}}

\newcommand{\our}{{$\Delta$-LFM}}

%% file: sec/0_abstract.tex
\definecolor{darkpink}{RGB}{200, 50, 100}

\begin{abstract}

Understanding disease progression is a central clinical challenge with direct implications for early diagnosis and personalized treatment. While recent generative approaches have attempted to model progression, key mismatches remain: disease dynamics are inherently continuous and monotonic, yet latent representations are often scattered, lacking semantic structure, and diffusion-based models disrupt continuity with a random denoising process. In this work, we propose to treat the disease dynamics as a velocity field and leverage Flow Matching (FM) to align the temporal evolution of patient data.  Unlike prior methods, it captures the intrinsic dynamics of disease, making the progression more interpretable. 
However, a key challenge remains: in latent space, autoencoders (AEs) do not guarantee alignment across patients or correlation with clinical-severity indicators (\eg~age and disease conditions). 
To address this, we propose to learn patient-specific latent alignment, which enforces patient trajectories to lie along a specific axis, with magnitude increasing monotonically with disease severity. This leads to a consistent and semantically meaningful latent space. Together, we present \our, a framework for modeling patient-specific latent progression with flow matching. Across three longitudinal MRI benchmarks, \our~demonstrates strong empirical performance and, more importantly, offers a new framework for interpreting and visualizing disease dynamics. Code available: \href{https://github.com/chqwer2/Delta-LDM-Longitudinal}{\textcolor{darkpink}{https://github.com/chqwer2/Delta-LDM-Longitudinal}}.

% Flowing Latents

% Towards Interpretable Disease Progression via Latent Flow Matching

% Learning Disease Trajectories as Flows in Patient-Specific Latent Space

% Synthesizing the progression of degenerative diseases via medical imaging involves more than simply creating visually realistic images. It requires accurately capturing the intricate anatomy of the human body and modeling how it evolves over time. Although recent advances in image generation have achieved impressive results in producing photo-realistic outputs that appeal to the human eye, applying these techniques to medical imaging continues to present major challenges. Crucially, current approaches lack fine-grained, patient-specific anatomical control and struggle to model the nuanced trajectories of disease progression, both of which are fundamental for producing clinically meaningful and trustworthy outcomes. 
% In this work, we seek to explore these critical challenges in longitudinal generation by  proposing two key components. 
% To mitigate geometric inconsistencies, we introduce a graph-based anatomical prior that preserves structural fidelity and enables controlled synthesis. Furthermore, we propose a patient-specific longitudinal prior, learned through contrastive alignment, to accurately guide individual disease progression synthsis over time.  We integrate these two components into a unified framework, which we term Anovus. Comprehensive experiments and evaluations demonstrate the effectiveness of our method in achieving both anatomical fidelity and temporal consistency. 

\end{abstract}

%% file: sec/1_intro.tex
\section{Introduction}
\label{sec:intro}

Modeling disease progression is a fundamental problem in healthcare~\citep{suk2013deep,wang2020deep}, with profound implications for early detection and individualized treatment. Disease does not evolve uniformly: differences in anatomy, genetics, and environment give rise to heterogeneous trajectories, which are an intrinsic property of disease dynamics~\citep{real2007spatial,parratt2016infectious,li2025pants}. Accounting for this heterogeneity is essential for accurately characterizing temporal changes, identifying preclinical biomarkers, and designing personalized therapies that can improve clinical outcomes~\citep{frisoni2010clinical}. 
However, most progression models capture only population-level trends, overlooking individualized variation~\citep{rokuss2025lesionlocator,yang2025medical}. This limitation obscures early-stage signals and misaligns disease severity with underlying anatomy, ultimately reducing clinical utility. Capturing patient-specific dynamics is therefore critical, but remains an underexplored challenge~\citep{lai2025patient}.

Recent advances in generative modeling have opened new opportunities for studying disease progression. Traditional biomarker-based approaches often reduce complexity to coarse measures (\eg~ volume changes~\citep{jack2003mri,tabrizi2011biological} or disease rating~\citep{o2008staging,young2020imaging}), which can obscure heterogeneous disease dynamics. In contrast, longitudinal medical image generation offers a principled framework for visualizing symptomatic transitions and individualized patterns of anatomical change. Rather than limiting analysis to numeric outcomes, synthesized images yield interpretable and more informative visual representations that support structured analyses and offer more actionable insights for clinical practice~\citep{wu2024freetumor,chen2024analyzing,chen2025scaling}.

% presymptomatic

Several methods have attempted to model disease progression. For instance, DiffuseMorph~\citep{kim2022diffusemorph} introduces a morph field to describe pixel-wise displacements, and TADM~\citep{litrico2024tadm} incorporates temporal awareness by predicting residual images conditioned on age gaps. Despite these advances, generating temporally consistent and individualized disease trajectories remains an open problem. Beyond high-fidelity image synthesis, it is crucial for generative models to preserve fine-grained anatomical structures and accurately capture patient-specific dynamics. Although BrLP~\citep{puglisi2024enhancing} takes a step toward personalization by modeling volume changes, such guidance is coarse and provides only limited control over individualized trajectories.

\paragraph{Contributions: } 
Motivated by these gaps, we propose a personalized generative framework for modeling disease progression based on flow matching. Our approach, termed Progression Latent Flow Matching (\our), is designed to generate trajectories at arbitrary future time points within a patient-specific latent space, ensuring temporal consistency and clinical interpretability. The key contributions are summarized as follows:

\begin{itemize}[leftmargin=*]
    \item  We reformulate conventional flow matching, which typically models dynamics from $t=0$ to $t=1$, into a formulation where the time variable $t$ explicitly encodes the future time gap. This extension transforms the horizon from a normalized  range $[0,1]$ to a  meaningful temporal range $[0,T]$, enabling prediction at arbitrary future time points $T$ with consistent temporal semantics.

    \item We introduce a patient-specific latent space, regularized by a new \textit{ArcRank} loss, which enforces chronology-aware alignment and captures individualized disease dynamics.  
    
    \item We show that the learned latent space supports interpretable visualization of patient trajectories. Notably, although disease severity is not used for training, the latent representations naturally reflect severity levels, providing clinically meaningful structure.  
    
    \item We further  validate \our~on longitudinal MRI benchmarks and propose a progression-specific evaluation metric tailored to disease modeling called $\Delta$-RMAE. Results demonstrate improved imaging fidelity and more accurate alignment with actual disease progression.
    %, and enhanced interpretability compared to existing approaches.  

\end{itemize}

%% file: sec/2_related.tex
\section{Related Work}
\label{sec:related}

\noindent\textbf{Generative models for disease progression.}
Generative modeling has emerged as a powerful strategy for simulating longitudinal disease changes. Early efforts relied on generative adversarial networks (GANs), either by introducing morphological priors to simulate aging effects~\citep{raviDegenerativeAdversarialNeuroImage2019a} or by predicting deformation fields with 3D conditional GANs instead of direct intensity values~\citep{ravi2022degenerative}.

More recently, diffusion models~\citep{ddpm} have become the dominant paradigm due to their superior fidelity and stability. Extensions of this framework introduce temporal awareness and anatomical constraints. Sequence-Aware Diffusion Model (SADM) leverages multiple scans to autoregressively generate future MRIs~\citep{yoon2023sadm}, while DiffuseMorph~\citep{kim2022diffusemorph} produces smooth voxel-wise deformation fields across timepoints, thereby preserving topology. 
Progressive Image Editing (PIE)~\citep{liang2023pie} edits disease-related regions in medical images using text-guided diffusion models to reflect how a patient’s condition evolves over time. Kyung~\etal~\citep{kyung2024towards} adopt electronic health records (EHRs), including medical history, to model temporal changes and enable medically meaningful generation. 
Temporally-Aware Diffusion Model (TADM) conditions on age maps and adopts a residual prediction strategy, modeling progression as incremental changes rather than absolute reconstructions~\citep{litrico2024tadm}.

Conditioning strategies further enhance subject specificity and anatomical fidelity. Brain Latent Progression (BrLP) employs ControlNet with volumetric ratio conditioning to synthesize individualized disease trajectories while integrating population priors~\citep{puglisi2024enhancing, puglisi2025brain}. Similarly, BrainMRDiff fuses multiple structural masks (\eg~white matter, gray matter, ventricles) into a unified conditioning representation, substantially improving anatomical realism~\citep{bhattacharya2025brainmrdiff}. 
ImageFlowNet~\citep{liu2025imageflownet} learns multiscale joint patient embeddings and position-parameterized neural flow fields to forecast full image-level trajectories.  

% Islam~\etal~\citep{islam2025longitudinalflowmatchingtrajectory} propose a flow-matching model that  builds local probabilistic interpolants between successive timepoints. 

In contrast to these approaches, which primarily focus on realism and population-informed priors, our method targets patient-specific disease trajectories while explicitly modeling heterogeneity through a flow-matching generative framework.

\noindent\textbf{Subject-level consistency and biomedical plausibility.}
Ensuring subject-specific consistency and biomedical plausibility is a central challenge in modeling synthetic disease progression. Early approaches such as the Longitudinal Variational Autoencoder~\citep{ramchandran2021longitudinalvariationalautoencoder} and Longitudinal Self-Supervised Learning (LNE)~\citep{zhao2021longitudinal} focus on learning patient-temporal dynamics by constructing temporally coherent latent spaces, thereby producing individualized representations for downstream tasks.

Subsequent works introduced explicit mechanisms to regularize longitudinal consistency. Ren \etal~\citep{ren2023localspatiotemporalrepresentationlearning} enforce subject-level coherence through local and multi-scale spatio-temporal features within a self-supervised framework. In a complementary direction, Treatment-Aware Diffusion (TaDiff)~\citep{liu2025treatment} incorporates prior scans and treatment information to jointly generate future MRIs and tumor segmentations, thereby improving biological plausibility by conditioning on clinically relevant variables.

% Beyond modeling strategies, recent efforts have emphasized evaluation of biomedical plausibility. Yassin \etal~\citep{pmlr-v250-yassin24a} assess how well synthetic MRIs replicate Alzheimer’s disease and age-related changes, not only through visual inspection but also by quantifying structural hallmarks such as hippocampal atrophy and ventricular enlargement. 

%% file: sec/3_method.tex
\begin{figure}
    \centering
    \includegraphics[width=1.0\linewidth]{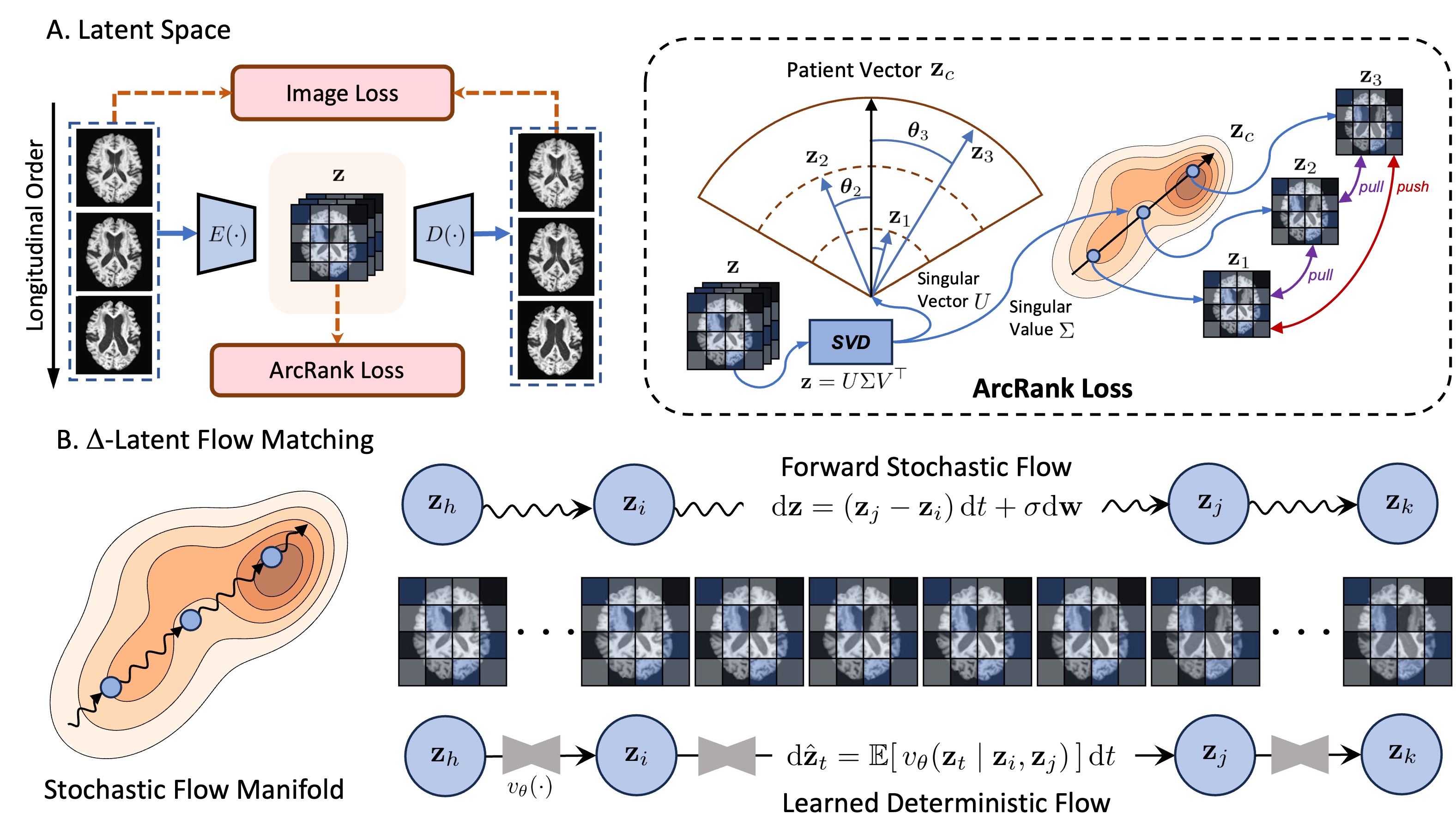}
    \caption{\our~overview. \our~operates in the latent space. In the first stage, an autoencoder constructs the latent space with ArcRank Loss to capture patient-specific disease trajectories. In the second stage, flow matching predicts disease progression along the trajectory over the total time horizon $T/\text{d}t$, where, for example, $T = j - i$,  to adapt the actual prediction span rather than relying on a fixed interval.}
    \label{fig:placeholder}
\end{figure}

\section{Methodology}
\label{sec:method}

Our proposed approach consists of two key components:  
(i) patient-specific latent learning via \textit{ArcRank} (Sec.~\ref{sec:arcrank}), and  
(ii) arbitrary-time progression modeling via {\our} (Sec.~\ref{sec:delta}).  

Before delving into these components, we first present the preliminaries in Sec.~\ref{sec:prelim}.

\subsection{Preliminaries}
\label{sec:prelim}

Flow matching~\citep{lipman2023flow, tong2023improving} trains continuous normalizing flows by directly aligning vector fields instead of relying on stochastic denoising objectives as in diffusion models. 
Let $\pi_0$ and $\pi_1$ denote the source and target distributions, respectively. Flow matching defines a family of intermediate distributions $\{\pi_t\}_{t \in [0,1]}$ along a continuous trajectory connecting $\pi_0$ and $\pi_1$. The goal is to learn a time-dependent velocity field $v_\theta(\mathbf x,t)$ such that the solution of the ordinary differential equation (ODE)~\citep{lu2022dpm} transports $\pi_0$ exactly to $\pi_1$ at $t=1$:
\begin{equation}
    \frac{d \mathbf x_t}{d t} = v_\theta(\mathbf x_t, t),  %\quad  x_t = (1-t)x_0 + t x_1, 
    % \quad x_t \sim \pi_0, %\quad  x_1 \sim \pi_1,
\end{equation}
 where the velocity field $v_\theta(\mathbf x,t)$ is trained to ensure that  samples from $\pi_0$ flow continuously into $\pi_1$.

Unlike score-based diffusion models, which approximate the score function $\nabla_{\mathbf x} \log \pi_t(\mathbf x)$ via noisy denoising objectives, flow matching bypasses score estimation by directly regressing the neural vector field $v_\theta(\mathbf x,t)$ onto analytically known target fields. For example, under conditional flow matching~\citep{lipman2023flow}, one defines a reference path $\mathbf x_t = (1-t)\mathbf x_0 + t \mathbf x_1$ between samples $\mathbf x_0 \sim \pi_0$ and $\mathbf x_1 \sim \pi_1$, yielding a target velocity
\begin{equation}
    v^\star(\mathbf x_t, t) = \mathbf x_1 - \mathbf x_0,
\end{equation}
which is constant along the trajectory. The training objective then minimizes
\begin{equation}
    \mathcal{L}_\text{FM}(\theta) = \mathbb{E}_{\mathbf x_0 \sim \pi_0, \, \mathbf x_1 \sim \pi_1, \, t \sim \mathcal{U}[0,1]} 
    \big\| v_\theta(\mathbf x_t, t) - v^\star(\mathbf x_t, t) \big\|^2,
\end{equation}
which yields stable training and ensures the learned flow recovers the correct marginal $\pi_1$ at $t=1$.

To further incorporate stochasticity~\citep{albergo2023stochastic}, the deterministic flow can be extended 
to a stochastic differential equation of the form
\begin{equation}
    \mathrm{d}\mathbf{z}_t 
    = v^\star(\mathbf{z}_t, t)\,\mathrm{d}t 
    + \sigma \,\mathrm{d}\mathbf{w}_t,
\end{equation}
where $\mathbf{w}_t$ is a Wiener process and $\sigma$ controls the noise scale. 
This extension allows flow matching to model both deterministic transport and 
stochastic variability, providing greater flexibility in capturing complex 
data distributions.
% This formulation yields stable training dynamics 

\subsection{ArcRank: Longitudinal Alignment of Patient Latent}
\label{sec:arcrank}
% \subsubsection{Patient-specific Latent Space} 
 To enhance sensitivity to patient-specific dynamics, we introduce a chronologically-aware latent representation learned via a new contrastive objective, termed \textit{ArcRank Loss}. This objective aligns representations across time for the same individual by enforcing both angular consistency and temporal ordering in the embedding space. Let $\mathbf{z}$ denote the latent representation of an input $\mathbf{x}$, extracted from a Variational Autoencoder (VAE)~\citep{kingma2013auto} encoder $E(\cdot)$. The loss encourages
\begin{equation}
    \angle(\mathbf{z}_i) \approx \angle(\mathbf{z}_j) 
    \quad \land \quad 
    \|\mathbf{z}_i\| \prec \|\mathbf{z}_j\|
    \quad \text{if } t_i < t_j,
    \qquad 
    \mathbf{z}_i, \mathbf{z}_j \in \mathcal{P}^k,
\end{equation}
where $\mathcal{P}^k$ denotes the latent feature set of patient $k$, and $t$  here represents the imaging capture time. Here, $\angle(\cdot)$ represents the direction of a latent vector, and $\|\cdot\|$ is its magnitude. Intuitively, samples from the same patient are encouraged to align along a consistent latent direction, while the trajectory respects chronological order (further along the trajectory corresponds to later disease stages).

In practice, we extract these components via Singular Value Decomposition (SVD)~\citep{klema1980singular} and write
\begin{equation} \angle \mathbf{z} = U, \quad \|\mathbf{z}\| = \Sigma, \quad \text{where } \; U \Sigma V^\top = \operatorname{SVD}(\mathbf{z}). \end{equation}
where $U$ captures the orientation (angle) and $\Sigma$ encodes the scaling (norm).

% where $U$ captures the orientation (angular information) and $\Sigma$ encodes the scaling (norm-related information).

% The proposed \textit{ArcRank Loss} is designed to capture both directional consistency and temporal progression in patient-specific feature trajectories. 

\paragraph{ArcRank Loss Computation.}  
For a given patient $k$, let $\mathcal{P}^k = \{\mathbf{z}_1, \mathbf{z}_2, \cdots,\mathbf{z}_n\}$ denote latent features at times $t_1 < t_2 < \cdots< t_n$. We decompose each feature via SVD:
\begin{equation}
    \angle(\mathbf{z}_{t}) = U_t, \qquad \|\mathbf{z}_{t}\| = \Sigma_t.
\end{equation}
The angular consistency loss encourages stability of feature orientations across time, while the temporal ranking loss enforces monotonic growth of feature magnitudes:
\begin{equation}
\mathcal{L}_{\text{Arc}} = 
\sum_{\substack{i < j \\ \text{same patient}}} 
\lvert U_i - U_j \rvert,
\quad
\mathcal{L}_{\text{Rank}} = 
\sum_{\substack{i < j \\ \text{same patient}}} 
\max\big(0,\, m - (\Sigma_j - \Sigma_i)\big),
\quad t_i < t_j,
\end{equation}
where $m > 0$ is a margin hyperparameter. 
The final objective combines the two components:
\begin{equation}
\mathcal{L}_{\text{ArcRank}} 
= \lambda_{\text{arc}} \, \mathcal{L}_{\text{Arc}}
+ \lambda_{\text{rank}} \, \mathcal{L}_{\text{Rank}},
\end{equation}
where $\lambda_{\text{arc}}, \lambda_{\text{rank}} > 0$ balance the contributions.

In practice, the ranking term $\mathcal{L}_{\text{Rank}}$ tends to push timepoints for the same patient increasingly far apart, and the margin $m$ is shared across all pairs $(i,j)$, i.e., it is not adaptively optimized for different temporal gaps. To mitigate this effect, we introduce an additional pull term that penalizes excessive separation in magnitude between temporally adjacent features:
\begin{equation}
    \mathcal{L}_{\text{Pull}} 
    =  \big\lvert \Sigma_{j} - \Sigma_i \big\rvert.
\end{equation}
We augment the ranking objective as
\begin{equation}
    \tilde{\mathcal{L}}_{\text{Rank}} 
    = \mathcal{L}_{\text{Rank}} + \mathcal{L}_{\text{Pull}},
\end{equation}
using the same overall weight as $\mathcal{L}_{\text{Rank}}$ in the total loss. This pull term softly encourages consecutive timepoints to remain close in latent space, while still allowing the hinge-based push term to enforce the desired temporal ordering. 

To stabilize training, for each pair $(i,j)$ used in the ArcRank loss, we apply the stop-gradient operator $\operatorname{sg}(\cdot)$ to the representation of $i$.

% Here we provide the AE reconstruction results when using the pull term versus not using it:

% | Ablation | PSNR | SSIM |
% | -------- | -------- | -------- |
% | w/ Pull     | 32.49   |  94.71     |
% | w/o Pull      | 30.73    | 91.84    |

% We did not evaluate the Flow Matching results in the "without-pull" setting because the reconstruction baseline is already too low. 

\subsection{\our: Flow Matching along Patient Trajectories}
\label{sec:delta}

While ArcRank enforces patient-specific chronological alignment, it does not explicitly capture how latent states evolve over time. To address this limitation, we introduce \our, a flow-matching objective that learns smooth mappings across arbitrary time intervals within a patient’s trajectory. Unlike standard flow matching, which interpolates between fixed source and target distributions (\eg~from $0 \to 1$), \our~models temporal transitions on disease-specific time scales $T$, where $T$ denotes the future time relative to the baseline. For clarity, we refer to the conventional $[0,1]$ interval as \textit{Physical Sampling}, and to our proposed $[0,T]$ formulation as \textit{Temporal Sampling}.

Formally, given a patient $k$, let $<\mathbf{z}_i,\ \mathbf{z}_j>$ denote paired latent states at times $t_i < t_j$. 
We define the latent transition field ${v}_\theta(\cdot)$, parameterized by a neural network, which approximates the continuous-time velocity of latent dynamics. The goal is to match this learned velocity field to the empirical difference between latent features across time:
\begin{equation}
v^\star(i,j) 
\;\approx\; \frac{\mathbf{z}_j - \mathbf{z}_i}{t_j - t_i}
,
\quad t_j > t_i, 
\end{equation}
The associated flow-matching loss is then given by
\begin{equation}
\mathcal{L}_{\text{LFM}}(\theta)
= \sum_{i<j}
\Big| {v}_\theta(i,j) 
- v^\star(i,j) \Big|^2.
\end{equation}

\paragraph{Inference.} 
Given an input state $\mathbf{x}_i$ observed at time $t_i$, the encoder maps it to a latent representation $\mathbf{z}_i$. To predict the target state $\mathbf{x}_j$ at a later time $t_j$, \our~integrates the learned velocity field forward in time. Specifically, we fix a discretization step size $\text dt > 0$ (\eg~$\text dt = 0.01$), so that the total number of integration steps is
\begin{equation}
N = \frac{t_j - t_i}{\text dt}.
\end{equation}

At each step, the latent state is updated according to the learned dynamics
\begin{equation}
\mathbf{z}_{i+\text dt} \;=\; \mathbf{z}_{i} \;+\; \text dt \cdot v_\theta(\mathbf{z}_i, t_i),
\end{equation}
with time advanced as $t_{i+\text dt} = t_i + \text dt$. Repeated application of this update yields an approximation of the latent state at $t_j$, which is then decoded to produce the predicted observation $\hat{\mathbf{x}}_j$.

\subsection{Discussion}

\paragraph{The use of Flow Matching.}   
We use flow matching because it naturally models continuous trajectories while keeping intermediate states interpretable, which is crucial in our setting. It learns a continuous velocity field between observed endpoints and allows direct sampling of intermediate states, without relying on long numerical integrations. Compared to latent ODEs, this avoids sensitivity to ODE solver choices and reduces computational cost. While rectified flow methods are effective, they are typically built on diffusion trajectories; in contrast, flow matching gives a more direct and tractable formulation of the probability path that is better aligned with our goal of faithful temporal interpolation.

\paragraph{Latent-space trajectories reasoning.}  
One central assumption in our framework is that disease progression can be represented as a straight line in latent space, structured along a  linear trajectory where the magnitude encodes disease stage and the direction preserves patient identity. While this design simplifies prediction, it raises an important question: can a straight line adequately capture both the shared and divergent aspects of disease dynamics across patients? We argue that this concern is valid. In our work, however, the linear trajectory is not intended to reproduce the full biological complexity of disease, but rather to provide a stable and interpretable scaffold that highlights the primary mode of variation corresponding to disease advancement.

\paragraph{Temporal uneven progression.}  
We formulate flow matching (FM) over the interval $[0,T]$, interpreting the index as a progression coordinate rather than physical clock time. However, although this formulation accounts for time differences, it implicitly assumes that progression is evenly distributed along the disease trajectory. In practice, progression is often uneven: extended periods of stability may be punctuated by sudden changes, or vice versa. The constant-velocity assumption in FM fails to capture this heterogeneity, effectively treating patient scans as uniformly spaced along the trajectory. To address this limitation, we introduce conditioning on the target time $T$ together with patient-specific attributes (\eg~sex, age, and clinical status) to enable the model to capture heterogeneous progression patterns that vary substantially over time.

\paragraph{Design details of ArcRank loss.}  
The proposed ArcRank loss uses SVD to jointly capture the direction and magnitude of latent trajectories. Alternatives such as cosine similarity for angles and absolute values for magnitudes treat these aspects separately and can become unstable when latent features are noisy or vary in scale. 
In contrast, SVD provides an orthogonal basis in which singular vectors naturally describe trajectory directions and singular values represent their magnitudes. This unified representation is compact and numerically stable, allowing ArcRank loss to align trajectories along meaningful progression directions while penalizing inconsistent scaling. 

%% file: sec/4_experiment.tex
\section{Experiments}
\label{sec:exp}

\begin{figure}[t]
    \centering
    \includegraphics[width=1\linewidth]{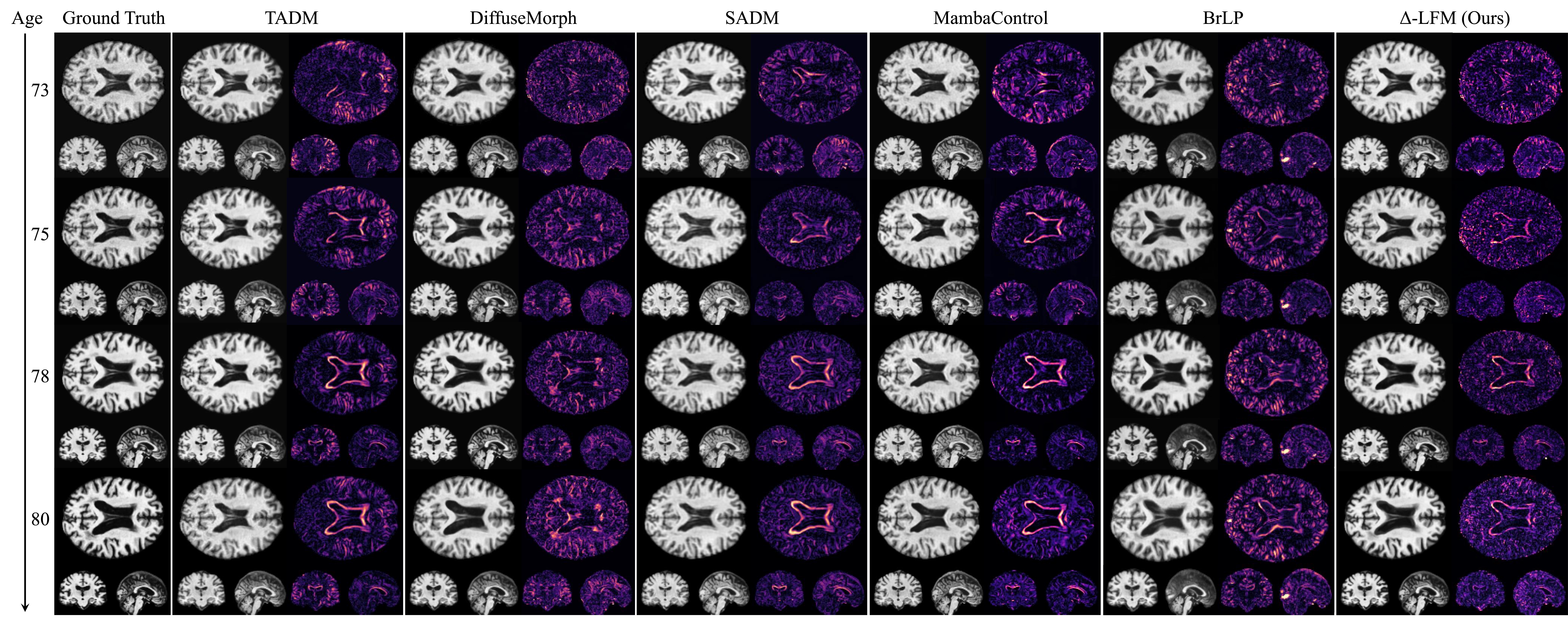}
    \caption{\textbf{Method comparison.} The first column shows the ground truth. All predictions are initialized from the same baseline scan at age 71. Odd columns show predicted MRIs; even columns show residual maps relative to the ground truth. Residuals use the \texttt{magma} colormap, scaled to the 1$^{\text{st}}$ to 99$^{\text{th}}$ percentiles of the residual distribution, enhancing error visibility around the lateral ventricles (central butterfly-shaped region) and cortical gray matter (outer surface).}
    \label{fig:comparison}
\end{figure}

\subsection{Experiment Settings}

We build our study on three major Alzheimer’s disease (AD) cohorts: ADNI~\citep{petersen2010alzheimer}, OASIS-3~\citep{lamontagne2019oasis}, and AIBL~\citep{ellis2009australian}. Every MRI undergoes the same preprocessing pipeline: N4 bias correction~\citep{tustison2010n4itk}, skull stripping~\citep{hoopes2022synthstrip}, resampling to 1.5 mm$^3$ voxels. We then normalize tissue contrasts and register all scans to a refined standard brain  template. We use a random 80/5/15 split for train/validation/test.

We benchmark against a broad spectrum of recent approaches. Direct-prediction methods include CardiacAging~\citep{campello2022cardiac} and CounterSynth~\citep{pombo2023equitable}. Deformation-based modeling represented by DiffuseMorph~\citep{kim2022diffusemorph}. Diffusion-based methods span multiple variants: SADM~\citep{yoon2023sadm}, which denoises full images; TADM~\citep{litrico2024tadm}, which denoises residuals; and two ControlNet-based~\citep{zhang2023adding} models, BrLP~\citep{puglisi2024enhancing} and MambaControl~\citep{yang2025mambacontrol}.

\textbf{Metrics.} We evaluate reconstruction quality with two standard image-level metrics: Peak Signal-to-Noise Ratio (PSNR) and Structural Similarity Index Measure (SSIM). To capture anatomical fidelity, we followed  BrLP~\citep{puglisi2024enhancing} to report region-level mean absolute error (MAE) over clinically relevant structures (hippocampus, amygdala, lateral ventricles, cerebrospinal fluid (CSF), and thalamus).

However, measuring image similarity alone is insufficient in the longitudinal setting. Scans from the same patient naturally exhibit high similarity because they share the same biological structure. The subtle deformations caused by disease are often small and easily overshadowed by the stable, unaffected anatomy. As a result, conventional metrics such as PSNR and SSIM tend to report inflated scores, failing to reflect clinically meaningful progression.

The true signal of interest lies in the \textit{residual differences} between baseline and follow-up scans. These residuals often encode the disease trajectory. To capture this, we introduce a residual-based metric. Specifically, let 
\begin{equation}
    \Delta_{\text{gt}} = \mathbf{x}_T - \mathbf{x}_0, 
    \quad 
    \Delta_{\text{gen}} = \hat{\mathbf{x}}_T - \mathbf{x}_0,
\end{equation}
where $\Delta$ denotes the residual image, $\mathbf{x}$ and $\hat{\mathbf{x}}$ denote ground-truth and generated MRIs, respectively. The metric is defined as the Residual-based Relative Mean Absolute Error ($\Delta$-RMAE):
\begin{equation}
\Delta{\text{-RMAE}} 
= \frac{\big\lvert \Delta_{\text{gt}} - \Delta_{\text{gen}} \big\rvert}
{\tfrac{1}{2}\big(\lvert \Delta_{\text{gt}} \rvert + \lvert \Delta_{\text{gen}} \rvert\big)}.
\end{equation}
By definition, $\Delta$-RMAE $ \in [0,2]$. A smaller value indicates that the predicted residual aligns with the ground-truth residual, \ie~the model has correctly captured disease-relevant evolution. If the model predicts almost no change (essentially copying the baseline image), then $\Delta_{\text{gen}} \approx 0$. In this case, both numerator and denominator are dominated by $\lvert \Delta_{\text{gt}} \rvert$, and the score approaches 2. Likewise, if the prediction deviates strongly, $\lvert \Delta_{\text{gen}} \rvert$ dominates, again driving the score toward 2.

\input{table/01_comparison}

\subsection{Benchmark Comparisons}
We present qualitative results in Figure \ref{fig:comparison}. In this subject, disease progression primarily affects the lateral ventricles (the central butterfly-shaped region) and the cortical gray matter (outer surface). SADM, TADM, and MambaControl fail to capture changes in the lateral ventricles, resulting in pronounced butterfly-shaped error in the residual image. DiffuseMorph and BrLP better capture ventricular changes, but leave substantial residuals along the cortical surface. Our method also shows error in the ventricles; however, these are less intense, with thinner boundaries and smaller cortical-surface errors, demonstrating improved overall accuracy.

We report PSNR and SSIM in Table~\ref{tab:comparison_psnr}, and Region MAE together with $\Delta$-RMAE in Table~\ref{tab:comparison_mae}. Across all three datasets, our proposed $\Delta$-LFM consistently achieves high performance.

In Table~\ref{tab:comparison_psnr}, $\Delta$-LFM improves PSNR by a clear margin over the strongest baseline. On ADNI, it reaches 30.59 dB, exceeding the second-best method (MambaControl, 29.72 dB) by +0.87 dB. Similar gains are observed on AIBL (+0.66 dB) and OASIS (+0.77 dB). The SSIM results mirror this trend: $\Delta$-LFM achieves 94.62, 93.92, and 89.36, improving upon the best baselines by roughly +1.0, +0.7, and +1.1 points, respectively. These consistent improvements indicate that $\Delta$-LFM produces reconstructions with both higher fidelity and sharper structural alignment.

Table~\ref{tab:comparison_mae} further validates these observations with volumetric error metrics. $\Delta$-LFM yields the lowest Region MAE on ADNI (0.210) and OASIS (0.262), and matches the best-performing method (DiffuseMorph) on AIBL (0.226). More importantly, it achieves the lowest $\Delta$-RMAE across all datasets: 0.436 on ADNI, 0.417 on AIBL, and 0.473 on OASIS. Compared to the strongest baseline (MambaControl), this represents relative error reductions of about 21\%, 21\%, and 16\%, respectively.

\subsection{Ablation study}

\begin{figure}[t]
    \centering
    \begin{minipage}{0.48\linewidth}
        \centering
        % \includegraphics[width=\linewidth, trim=45 25 70 40, clip]{asset/tsne-vis/tsne_output_2D_brain.jpg}
        % \fbox{%
        % \includegraphics[width=\linewidth, trim=89.9 54.85 70 60, clip]{asset/tsne-vis/tsne_output_2D_brain.jpg}%
        \includegraphics[width=\linewidth, trim=90.05 54.8 70 60, clip]{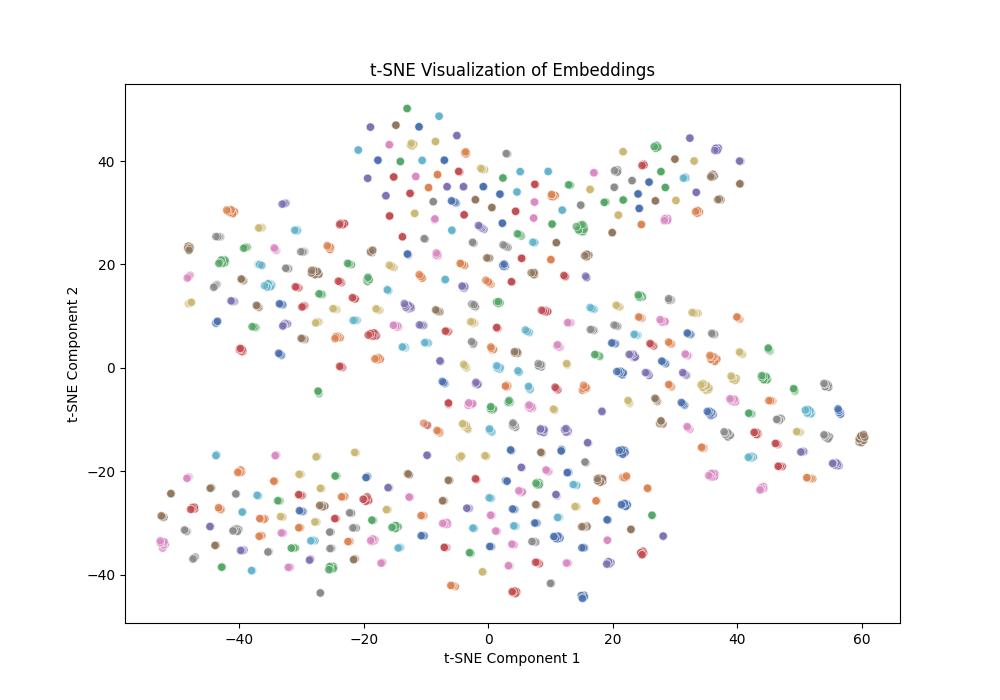}%
        % }
        \caption*{(a) colored by patient identity}
    \end{minipage}
    \hfill
    \begin{minipage}{0.48\linewidth}
        \centering
        \includegraphics[width=\linewidth, trim=77.6 47.2 60 49, clip]{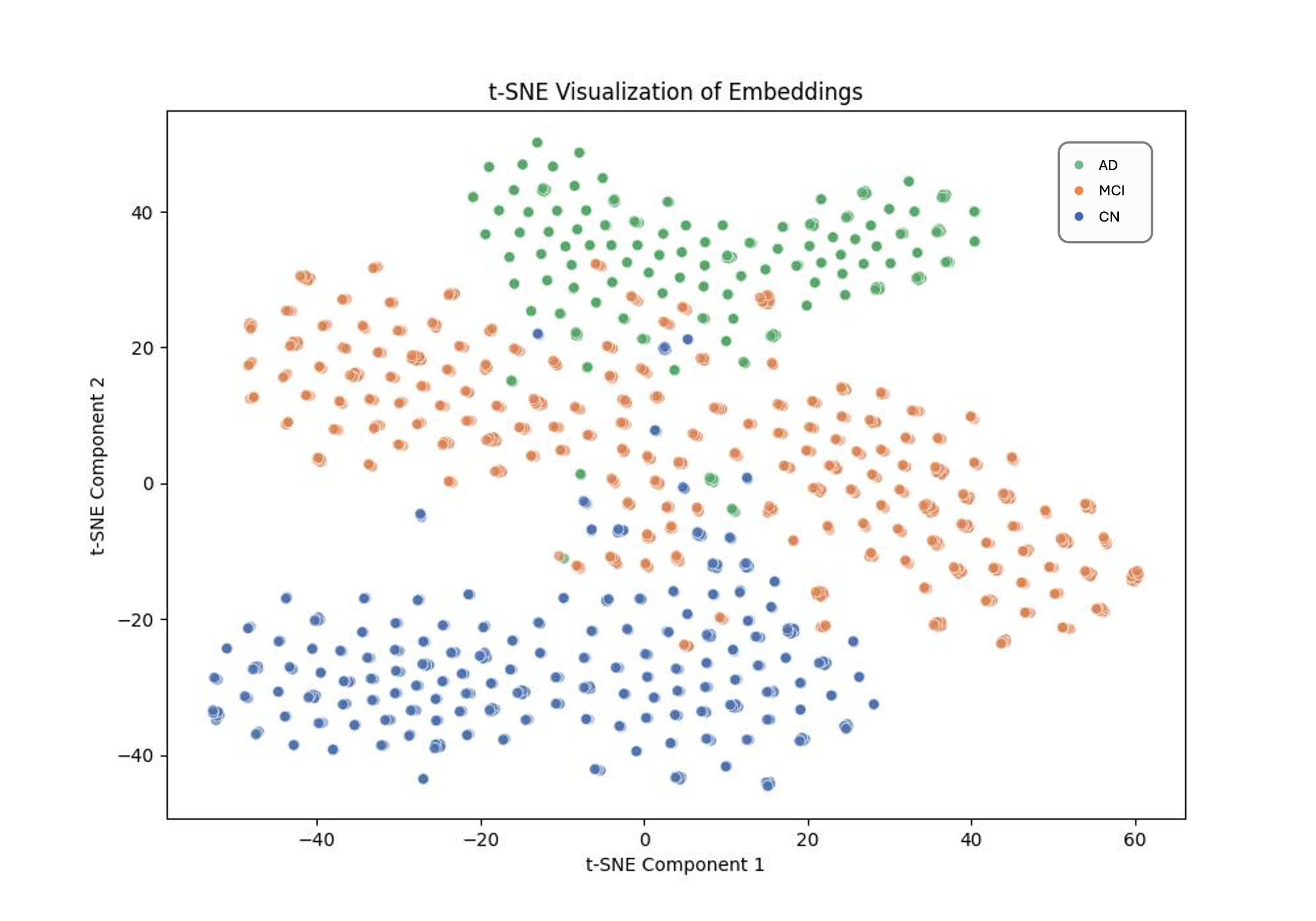}
        \caption*{(b) colored by diagnosis status}
    \end{minipage}

\caption{t-SNE projection of the learned longitudinal latent space into two dimensions. 
(a) Coloring by subject ID shows that scans from the same patient cluster together. 
(b) Coloring by disease status shows that scans with the same diagnosis are grouped closely.  
CN: Cognitively Normal. MCI: Mild Cognitive Impairment. AD: Alzheimer’s Disease (Dementia Stage).
% \review{Removing top and right borders and enlarging the labels and tick labels.}
}
    \label{fig:tsne_latent_space}
\vspace{-5pt}
\end{figure}

We provide visualization results of the learned latent space in Figure~\ref{fig:tsne_latent_space} on test cases in ADNI dataset, where t-SNE is used to project the representations into two dimensions. As shown in the left panel, the latent trajectories of individual patients are well aligned. Interestingly, the right panel reveals that the latent space also organizes according to diagnostic status, despite the fact that this information was never used during training.

\input{table/02_ablation}

To assess the contribution of each component in our framework, we conducted an ablation study across three datasets, as summarized in Table~\ref{tab:ablation}. The experiments examine the effect of (i) introducing conditional information, (ii) different auto-encoder training strategies (Arc Loss, Rank Loss, ArcRank Loss), and (iii) different flow-matching (FM) manifold sampling strategies, namely physical sampling $[0,1]$ versus temporal progression modeling $[0,T]$.

\input{table/03_combine_ablation}

The baseline Latent Flow Matching (LFM)~\citep{dao2023flow} model without conditioning performs the worst across all metrics, showing notably low SSIM/PSNR and high error values. Incorporating conditional information (Conditional LFM) significantly improves image quality (SSIM from 88.62 to 91.20; PSNR from 27.59 to 28.46) and reduces both Region MAE and $\Delta$-RMAE.

Within the $\Delta$-LFM ablation, we observe two consistent trends. First, the choice of loss function has a direct impact: using Arc Loss alone yields clear improvements, whereas Rank Loss alone slightly undermines performance. This is because Arc Loss encourages patient trajectories to follow a meaningful direction, while Rank Loss without Arc imposes only a weak magnitude constraint. In contrast, their combination (ArcRank Loss) achieves the best balance, suggesting that structural alignment (Arc) and temporal ordering (Rank) are complementary. Second, extending the FM sampling strategy from $[0,1]$ to $[0,T]$ further improves results by promoting gradual residual modeling over meaningful temporal intervals.

\begin{figure}[t]
    \centering
    \includegraphics[width=1\linewidth]{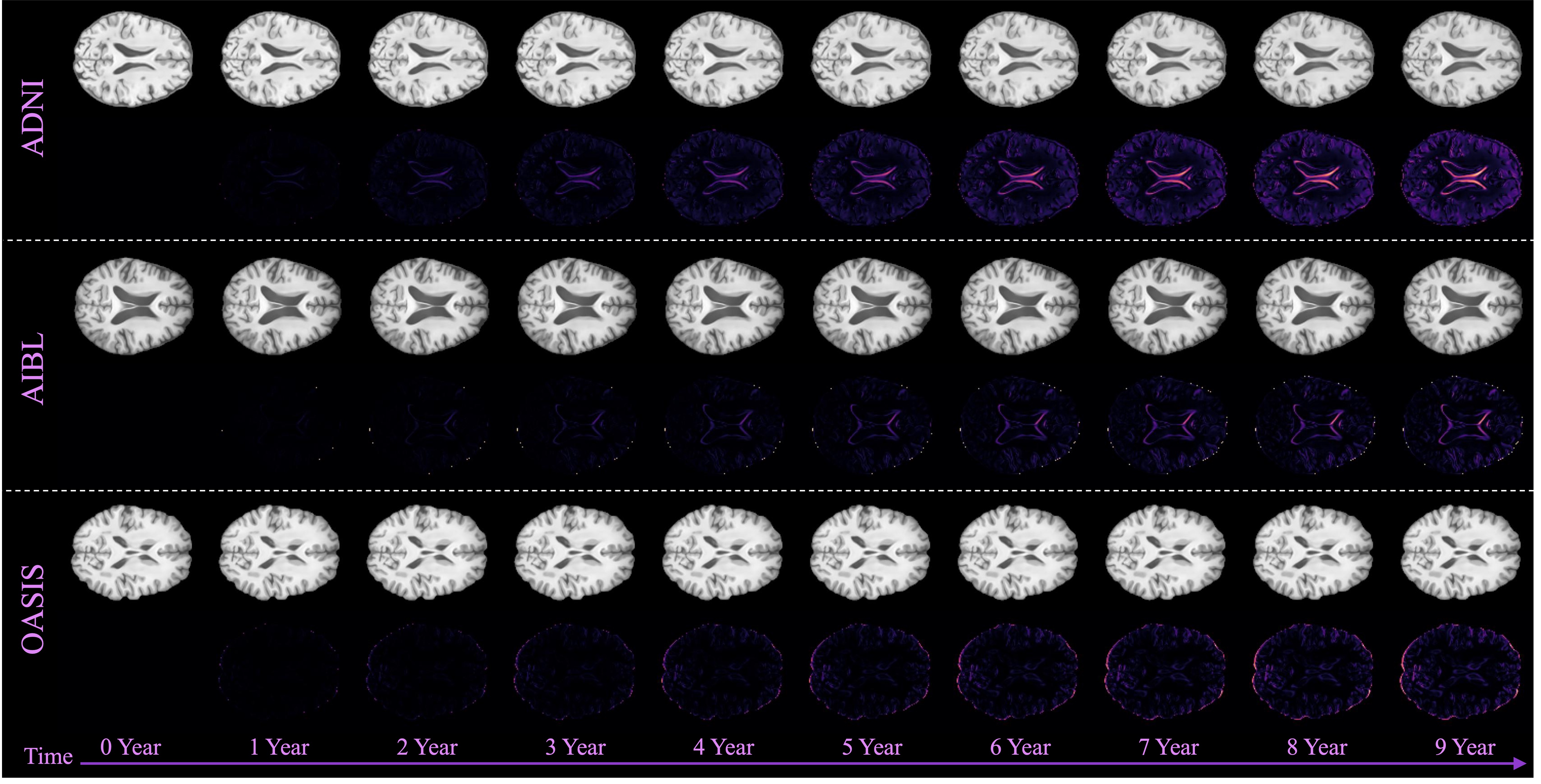}
     \caption{Continuous disease trajectory predicted by \our~at one-year intervals. Odd rows show the baseline MRI (year 0) and predicted MRIs from years 1–9; even rows show residual maps relative to baseline. Progression highlights structural abnormalities, primarily in atrophy-related regions (\eg~hippocampus, ventricles, surrounding cortex).}
    \label{fig:time_continuous}
\end{figure}

Finally, we visualize qualitative progression predictions in Figure~\ref{fig:time_continuous} using cases from three datasets, demonstrating the ability of \our~to predict future timepoints at arbitrary intervals. The results show disease-related deformations primarily affecting the ventricles and surrounding cortical areas, which is consistent with clinical observations.

We present qualitative visualizations of predicted disease progression in Figure~\ref{fig:time_continuous}. Each row illustrates disease trajectories over nine years generated by \our. The intermediate steps provide clinically interpretable insights into the progression process, capturing characteristic neurodegenerative patterns such as ventricular enlargement and cortical thinning in the temporal and parietal regions. These deformations align with established clinical observations, underscoring the biological plausibility of our model.

To further verify each proposed component, we provide additional ablations in Table~\ref{tab:combined_ablation}. We evaluate cosine similarity $\mathcal{L}_{\text{cosine}}$ as a surrogate for Arc loss and a simple ranking loss $\mathcal{L}_{\text{simple}}$ as a surrogate for Rank loss, defined as follows:
\begin{equation}
\mathcal{L}_{\text{cosine}}(i,j)
= 1 - \frac{\langle \mathbf{z}_i, \mathbf{z}_j \rangle}{\|\mathbf{z}_i\| \, \|\mathbf{z}_j\|}, 
\qquad
\mathcal{L}_{\text{simple}}(i,j)
= \max\bigl(0,\, d_j - d_i\bigr).
\end{equation}

Cosine similarity performs slightly worse than Arc loss, which is expected since it only enforces a global angular alignment and does not capture the fine-grained latent directional structure modeled by Arc loss. In contrast, the simple rank loss without a margin often leads to latent mode collapse, explaining its poor performance when combined with either Arc loss or cosine similarity.

\paragraph{Loss Weight.}

$\lambda_{\text{arc}}$ is relatively sensitive. Values $\lambda_{\text{arc}} > 0.01$ consistently cause collapse, while $0.0025$--$0.0075$ are stable across runs, with $\lambda_{\text{arc}} = 0.0075$ noticeably degrading AE reconstruction. We therefore set $\lambda_{\text{arc}} = 0.005$ as a compromise, trading a small decrease in AE reconstruction for improved flow-matching performance ($+0.4$ PSNR over $0.0025$).

For $\lambda_{\text{rank}}$, values above $0.05$ hinder convergence, whereas the range $0.005$--$0.02$ yields similar behavior. We choose $\lambda_{\text{rank}} = 0.01$, which gives a slight PSNR gain ($+0.05$). % without affecting AE reconstruction.

\paragraph{Speed Analysis}

ArcRank involves an SVD operation, which is computational expensive. In practice, we use 
\texttt{torch.linalg.svd(z, full\_matrices=False)}, which gives a $6\times$ speedup compared to the full matrices variant (0.0552s $\rightarrow$ 0.0092s on average over 50 runs). Overall, ArcRank increases AE training time by $\sim 40\%$ ($1.4\times$), which we regard as a reasonable overhead given the resulting performance gains.

%% file: table/01_comparison.tex
\begin{table}[t]
    \caption{Quantitative comparison of methods in terms of image quality metrics, including PSNR (dB) and \textbf{SSIM} ($\times 1.0 \mathrm{E}2$). Results are reported as mean $\pm$ standard deviation. \textit{Type} denotes the prediction strategy: ``Direct'' generates outputs directly, ``Deform'' estimates a transformation field, ``Noise'' applies diffusion-based denoising, ``$\Delta$Noise'' predicts and denoises progression residuals, and ``$\Delta$Flow'' models progression through gradual residual prediction.}
    \label{tab:comparison_psnr}
    \centering
    \setlength{\tabcolsep}{8pt}
    \renewcommand{\arraystretch}{1.3}
    \resizebox{\columnwidth}{!}{
    \begin{tabular}{l|c|cc|cc|cc}
    \toprule
   \multirow{2}{*}{\textbf{Method}} & \multirow{2}{*}{\textbf{Type}} 
  & \multicolumn{2}{c|}{\textbf{ADNI} } & \multicolumn{2}{c|}{\textbf{AIBL} } & \multicolumn{2}{c}{\textbf{OASIS} }\\ 
\cmidrule(lr){3-8}
 & & PSNR $\uparrow$ & SSIM  $\uparrow$ & PSNR $\uparrow$ & SSIM  $\uparrow$ & PSNR $\uparrow$ & SSIM  $\uparrow$   \\
\midrule

CardiacAging~\citep{campello2022cardiac}        & Direct        & 27.78 $\pm$ 1.49 & 92.04 $\pm$ 0.99 & 28.41 $\pm$ 1.41 & 90.30 $\pm$ 0.90 & 26.23 $\pm$ 1.45 & 86.17 $\pm$ 2.06 \\
CounterSynth~\citep{pombo2023equitable} & Direct & 28.24 $\pm$ 1.31 & 92.84 $\pm$ 0.95 & 28.96 $\pm$ 1.25 & 92.10 $\pm$ 0.91 & 26.79 $\pm$ 1.42 & 86.45 $\pm$ 1.98 \\
DiffuseMorph~\citep{kim2022diffusemorph}        & Deform        & 29.56 $\pm$ 1.63 & 93.57 $\pm$ 0.93 & 29.17 $\pm$ 1.54 & 92.70 $\pm$ 1.02 & 28.13 $\pm$ 1.36 & 88.16 $\pm$ 1.69 \\
SADM~\citep{yoon2023sadm}     & Noise         & 26.94 $\pm$ 2.28 & 85.15 $\pm$ 2.72 & 27.97 $\pm$ 1.91 & 89.40 $\pm$ 1.78 & 26.74 $\pm$ 1.80 & 86.36 $\pm$ 2.37 \\
TADM~\citep{litrico2024tadm}                & $\Delta$Noise & 27.89 $\pm$ 1.91 & 90.78 $\pm$ 1.54 & 28.58 $\pm$ 1.77 & 91.70 $\pm$ 1.01& 26.29 $\pm$ 1.52 & 85.51 $\pm$ 1.93 \\
ImageFlowNet~\citep{liu2025imageflownet} & Flow &  28.37 $\pm$ 1.18 	& 92.96 $\pm$ 0.88  & 29.08 $\pm$ 1.12 	& 92.23 $\pm$ 0.84 &  	27.92 $\pm$ 1.29   & 	87.63 $\pm$ 1.82 \\
BrLP~\citep{puglisi2024enhancing}                & Noise         & 28.51 $\pm$ 1.77 & 91.52 $\pm$ 1.31 & 28.96 $\pm$ 1.36 & 92.21 $\pm$ 0.97 & 27.98 $\pm$ 1.40 & 87.93 $\pm$ 1.44 \\
MambaControl~\citep{yang2025mambacontrol}        & Noise         & 29.72 $\pm$ 1.04 & 93.60 $\pm$ 0.96 & 29.86 $\pm$ 1.21 & 93.17 $\pm$ 0.98 & 28.24 $\pm$ 1.28 & 88.30 $\pm$ 1.52 \\
\midrule
$\Delta$-LFM (Ours)           & $\Delta$Flow    & \textbf{30.59 $\pm$ 0.89} & \textbf{94.62 $\pm$ 0.85} & \textbf{30.52 $\pm$ 1.03} & \textbf{93.92 $\pm$ 0.88} & \textbf{29.01 $\pm$ 1.19} & \textbf{89.36 $\pm$ 1.29} \\
\bottomrule
    \end{tabular}
    }
\end{table}

\begin{table}[t]
    \caption{Quantitative comparison results on  clinical structure faithfulness metrics are    reported in region-based MAE and $\Delta$-RMAE (Residual-based Relative Mean Absolute Error). Results are reported as mean $\pm$ standard deviation. }
    \label{tab:comparison_mae}
    \centering
    \setlength{\tabcolsep}{8pt}
    \renewcommand{\arraystretch}{1.3}
    \resizebox{\columnwidth}{!}{
    \begin{tabular}{l|c|cc|cc|cc}
    \toprule
   \multirow{2}{*}{\textbf{Method}} & \multirow{2}{*}{\textbf{Type}} 
  & \multicolumn{2}{c|}{\textbf{ADNI} } & \multicolumn{2}{c|}{\textbf{AIBL} } & \multicolumn{2}{c}{\textbf{OASIS} }\\ 
\cmidrule(lr){3-8}
 & & Region MAE $\downarrow$ & $\Delta$-RMAE $\downarrow$  & Region MAE $\downarrow$ & $\Delta$-RMAE $\downarrow$ & Region MAE $\downarrow$ & $\Delta$-RMAE $\downarrow$  \\
\midrule

CardiacAging~\citep{campello2022cardiac}        & Direct        & 0.289 $\pm$ 0.33 & 0.771 $\pm$ 0.12 & 0.281 $\pm$ 0.35 & 0.698 $\pm$ 0.13 & 0.345 $\pm$ 0.35 & 0.739 $\pm$ 0.13 \\
CounterSynth~\citep{pombo2023equitable} & Direct & 0.272 $\pm$ 0.32 & 0.704 $\pm$ 0.11 & 0.256 $\pm$ 0.30 & 0.672 $\pm$ 0.12 & 0.298 $\pm$ 0.34 & 0.690 $\pm$ 0.12 \\
DiffuseMorph~\citep{kim2022diffusemorph}        & Deform        & 0.230 $\pm$ 0.28 & 0.516 $\pm$ 0.10 & \textbf{0.226 $\pm$ 0.27} & 0.482 $\pm$ 0.10 & 0.279 $\pm$ 0.29 & 0.503 $\pm$ 0.11 \\
SADM~\citep{yoon2023sadm}     & Noise         & 0.283 $\pm$ 0.34 & 0.746 $\pm$ 0.11 & 0.278 $\pm$ 0.30 & 0.693 $\pm$ 0.12 & 0.365 $\pm$ 0.39 & 0.728 $\pm$ 0.12 \\
TADM~\citep{litrico2024tadm}                & $\Delta$Noise & 0.263 $\pm$ 0.32 & 0.697 $\pm$ 0.11 & 0.243 $\pm$ 0.30 & 0.654 $\pm$ 0.11 & 0.291 $\pm$ 0.34 & 0.685 $\pm$ 0.11 \\
ImageFlowNet~\citep{liu2025imageflownet} & Flow & 0.259 $\pm$ 0.30 & 0.589 $\pm$ 0.11 &  0.240 $\pm$ 0.29 &  0.561 $\pm$ 0.12 & 0.283 $\pm$ 0.32 &  	0.574 $\pm$ 0.12 \\
BrLP~\citep{puglisi2024enhancing}        & Noise         & 0.257 $\pm$ 0.32 & 0.630 $\pm$ 0.10 & 0.278 $\pm$ 0.31 & 0.594 $\pm$ 0.11 & 0.335 $\pm$ 0.36 & 0.622 $\pm$ 0.11 \\
MambaControl~\citep{yang2025mambacontrol}        & Noise         & 0.225 $\pm$ 0.30 & 0.554 $\pm$ 0.09 & 0.249 $\pm$ 0.30 & 0.525 $\pm$ 0.10 & 0.299 $\pm$ 0.32 & 0.561 $\pm$ 0.10 \\
\midrule
$\Delta$-LFM (Ours)           & $\Delta$Flow  & \textbf{0.210 $\pm$ 0.28} & \textbf{0.436 $\pm$ 0.08} & 0.226 $\pm$ 0.29 & \textbf{0.417 $\pm$ 0.10} & \textbf{0.262 $\pm$ 0.30} & \textbf{0.473 $\pm$ 0.08} \\
\bottomrule

    \end{tabular}
    }
\end{table}

%% file: table/02_ablation.tex
\begin{table}[t]
  \caption{Ablation study results (mean across datasets). We evaluate the effect of different auto-encoder training strategies (Arc Loss, Rank Loss, ArcRank Loss) and flow-matching manifold sampling strategies ($[0,1]$ vs.\ $[0,T]$). The Cond. represent using condition or not. The row with gray background indicated the proposed method. 
  }
    \label{tab:ablation}
    \centering
    \setlength{\tabcolsep}{8pt}
    \renewcommand{\arraystretch}{1.3}
    \resizebox{\columnwidth}{!}{
    \begin{tabular}{>{\centering\arraybackslash}p{3.6cm}|lcc|cc|cc}
    \toprule
\multirow{2}{*}{\textbf{Model}} &
\multirow{2}{*}{\textbf{Auto-Encoder}} &
\multirow{2}{*}{\textbf{FM Sampling}} &
\multirow{2}{*}{\textbf{Cond.}} &
\multicolumn{2}{c|}{\textbf{Image Quality}} &
\multicolumn{2}{c}{\textbf{Structure Fidelity}} \\
\cmidrule(lr){5-6}\cmidrule(lr){7-8}
& & & &
PSNR $\uparrow$ &  SSIM $\uparrow$ &
Region MAE $\downarrow$ &
$\Delta$-RMAE $\downarrow$ \\
\midrule
 LFM (Baseline)   & - & $[0,\,\,1]$ &  \ding{55} & 27.59 $\pm$ 1.78 & 88.62 $\pm$ 1.63 & 0.357 $\pm$ 0.342 & 0.552 $\pm$ 0.146 \\
 Conditional LFM   & -& $[0,\,\,1]$ & \ding{51}   & 28.46 $\pm$ 1.22 & 91.20 $\pm$ 1.15 & 0.299 $\pm$ 0.312 & 0.486 $\pm$ 0.101 \\
\midrule
  \multirow{5}{*}{\boldmath$\Delta$-LFM}
  & - & $[0,\,T]$   & \ding{51}  & 28.78 $\pm$ 1.15 & 90.97 $\pm$ 1.07 & 0.286 $\pm$ 0.295 & 0.472 $\pm$ 0.091 \\
  & w/ Arc Loss      & $[0,\,\,1]$ & \ding{51}   & 29.52 $\pm$ 0.98 & 92.38 $\pm$ 0.95 & 0.251 $\pm$ 0.282 & 0.457 $\pm$ 0.086 \\ 
  & w/ Rank Loss     & $[0,\,\,1]$ &\ding{51}    & 28.36 $\pm$ 1.21 & 91.15 $\pm$ 1.11 & 0.272 $\pm$ 0.285 & 0.474 $\pm$ 0.088 \\
  & w/ ArcRank Loss  & $[0,\,\,1]$ & \ding{51}   & 29.83 $\pm$ 0.95 & 92.74 $\pm$ 0.93 & 0.243 $\pm$ 0.277 & 0.454 $\pm$ 0.088 \\
\rowcolor{gray!15}  % << now valid (right after \\)
  & w/ ArcRank Loss  & $[0,\,T]$ & \ding{51}   & \textbf{30.04 $\pm$ 1.04} & \textbf{92.63 $\pm$ 1.03} & \textbf{0.233 $\pm$ 0.282} & \textbf{0.442 $\pm$ 0.087} \\
\bottomrule

\end{tabular}
    }
    \vspace{-2pt}
\end{table}

%% file: table/03_combine_ablation.tex
\begin{table}[t]
\centering
\vspace{-2pt}
\caption{Ablation on the proposed components.}
\vspace{-2pt}
\resizebox{0.65\columnwidth}{!}{
\begin{tabular}{l l c c c}
\toprule
\textbf{Angular loss} & \textbf{Ranking loss} & \textbf{Sampling} & \textbf{PSNR} & \textbf{SSIM} \\
\midrule
% Arc  Loss    & --      & $[0,\,\,1]$           & 29.52 & 92.38 \\
Arc   Loss   & --      & $[0,\,T]$           & 29.88 & 92.31 \\
% --       & Rank  Loss  & $[0,\,\,1]$           & 28.36 & 91.15 \\
--       & Rank  Loss  & $[0,\,T]$           & 28.85 & 91.10 \\
\midrule
Cosine  Similarity & --      &  $[0,\,\,1]$          & 29.13 & 91.10 \\
Cosine  Similarity  & Simple Rank Loss   &  $[0,\,\,1]$          & 28.32 & 89.92 \\
Arc    Loss  & Simple Rank Loss &  $[0,\,\,1]$          & 28.44 & 89.87 \\
% Arc    Loss   & Rank   Loss &   $[0,\,\,1]$         & 29.83 & 92.74 \\
\bottomrule
\end{tabular}
}
\label{tab:combined_ablation}
\vspace{-5pt}
\end{table}

%% file: sec/5_conclusion.tex
\section{Conclusion}
\label{sec:conclusion}

We introduce \textbf{Progression Latent Flow Matching (\our)}, a framework that unifies the proposed ArcRank loss with temporal flow matching to capture patient-specific disease trajectories. To evaluate progression in isolation, we propose a new metric, \(\Delta\)-RMAE. On longitudinal MRI benchmarks, \our~consistently achieves higher fidelity, lower error, and tighter alignment with ground-truth progression than prior methods. Beyond accuracy, it yields temporally coherent and clinically interpretable trajectories. These results show that \our~provides a new direction for progression modeling, with the potential to generalize across other temporal generation tasks.

%% file: sec/6_appendix.tex
\clearpage
\appendix

\par\vspace{2em}
{\LARGE Appendix\par}
\vspace{1em}

\section{Training Setting}
We adopt the AdamW optimizer~\citep{loshchilov2017decoupled} for all experiments. For the autoencoder (AE) and 3D U-Net~\citep{ronneberger2015u}, we trained with a learning rate of $1\times 10^{-3}$ and a batch size of $2$ for 300 epochs. For the flow matching model based on the 3D U-Net, we used a reduced learning rate of $3\times 10^{-5}$ with a larger batch size of $4$ for 200 epochs.

\section{Conditional Generation}

Following Litrico \etal~\citep{litrico2024tadm}, we encode the current sample time in flow matching into continuous embeddings using sinusoidal positional encodings. These time embeddings act as an additive bias to the input, ensuring that the generative process is aware of its temporal position.  

Building on this, and inspired by BrLP~\citep{puglisi2024enhancing}, we introduce \emph{explicit control signals} that combine patient-specific attributes with temporal queries. In particular, we condition the model not only on the start time, the current sample time, and the end query time, but also on the patient attributes. This design enables the model to incorporate both subject-specific and task-specific information when shaping the generative trajectory.  

To inject these conditional signals into the U-Net backbone, we explored three strategies:  (1) Additive biasing, (2) Cross-attention~\citep{vaswani2017attention} and (3) Adaptive Layer Normalization (AdaLN)~\citep{raffel2020exploring}.

Among these approaches, our experiments show that AdaLN is most effective  with our temporal sampling over the interval $[0, T]$. It adapts the network representation in a temporally aware and context-sensitive manner, allowing the generative process to be steered smoothly by patient attributes and target query times.

The implementation of AdaLN is realized through an initial one-layer MLP that processes the condition signals, followed by additional MLPs at each U-Net decoder layer to generate the corresponding AdaLN modulation parameters.

\section{Auto-encoder Ablation}

Table~\ref{tab:ae_ablation} summarizes the effect of AE capacity and training crop size. For the capacity ablation, we vary the encoder channels while fixing the crop size to $64^3$. For the crop-size ablation, we fix the AE to $[64,128,256]$ and compare $48^3$ vs.\ $64^3$ crops.

\begin{table}[h]
\centering
\caption{Ablation on AE capacity and training crop size.}
\resizebox{0.5\columnwidth}{!}{
\begin{tabular}{lccccc}
\toprule
\multicolumn{3}{c}{AE capacity (crop $64^3$)} & \multicolumn{3}{c}{Crop size (AE $[64,128,256]$)} \\
\cmidrule(lr){1-3}\cmidrule(lr){4-6}
\textbf{Channels} & \textbf{PSNR} & \textbf{SSIM} &
\textbf{Crop} & \textbf{PSNR} & \textbf{SSIM} \\
\midrule
$[64,128,256]$ & 30.04 & 92.63 & $64^3$ & 30.04 & 92.63 \\
$[64,128,128]$ & 29.89 & 92.55 & $48^3$ & 29.51 & 91.71 \\
$[64,128]$     & 29.62 & 92.10 &       &       &       \\
\bottomrule
\end{tabular}
}
\label{tab:ae_ablation}
\end{table}

We do not use deeper or wider AEs, as these configurations lead to out-of-memory (OOM) errors on our RTX A6000 GPU (48\,GB). Based on the above results, we use channels $[64,128,256]$ and crop size $64^3$ as our default configuration. Larger crops (e.g., $72^3$) also result in OOM on our hardware. To keep the comparison with baselines fair, we do not use a pretrained AE, although initializing with MAISI~\citep{guo2025maisi} can further improve PSNR (from 30.04 to 31.97 for $[64,128,256]$).

\section{Longitudinal Progression Quantitative Results}

Table~\ref{tab:progression_results} summarizes predictive performance across different future horizons. For short- to mid-term forecasts (1--5 years), reconstruction metrics remain consistently high: PSNR hovers around 31--32~dB and SSIM above 93\%, indicating strong fidelity in reproducing structural details. Region MAE is also relatively stable in this range ($\sim$ 0.20), suggesting reliable local accuracy, while $\Delta$-RMAE values remain below 0.41, reflecting well-preserved progression dynamics.  

Beyond 6 years, however, all metrics exhibit gradual degradation. PSNR and SSIM decline steadily, dropping to 28~dB and below 91\% by year~10, and further to 26.98~dB / 89.96\% at year~13. Concurrently, Region MAE and $\Delta$-RMAE increase monotonically, indicating that both absolute errors in regional values and their temporal progression deviate more strongly with longer horizons. Notably, the sharp rise in $\Delta$-RMAE after year~10 ($>0.50$) highlights a growing difficulty in capturing long-term progression trends.

We also observe that the prediction trend is not strictly linear as the prediction horizon increases.  
This is mainly because different time horizons are estimated from different subsets of subjects. 
In particular, fewer subjects have follow-up data at 10--13 years compared to 1--5 years, which introduces higher variability and heterogeneity, leading to slightly inconsistent trends.

\begin{table}
\centering
\caption{Quantitative results across different prediction horizons (years into the future). 
Measurements are reported using PSNR (dB), SSIM ($\times 1.0\text E2$), Region MAE, and \(\Delta\)-RMAE.}

\label{tab:progression_results}
\resizebox{0.8\textwidth}{!}{%
\begin{tabular}{>{\centering\arraybackslash}p{2cm}|
                >{\centering\arraybackslash}p{2.5cm}
                >{\centering\arraybackslash}p{2.5cm}
                >{\centering\arraybackslash}p{2.5cm}
                >{\centering\arraybackslash}p{2.5cm}}
\toprule
Year  & PSNR~\(\uparrow\) & SSIM~\(\uparrow\) & Region MAE~\(\downarrow\) & \(\Delta\)-RMAE~\(\downarrow\) \\
\midrule
1  & \(31.61 \pm 0.87\) & \(93.08 \pm 0.75\) & \(0.209 \pm 0.107\) & \(0.386 \pm 0.107\) \\
2  & \(31.79 \pm 0.92\) & \(93.79 \pm 0.77\) & \(0.217 \pm 0.106\) & \(0.369 \pm 0.097\) \\
3  & \(31.09 \pm 0.95\) & \(93.51 \pm 0.82\) & \(0.206 \pm 0.118\) & \(0.388 \pm 0.089\) \\
4  & \(32.48 \pm 0.93\) & \(93.95 \pm 0.89\) & \(0.201 \pm 0.132\) & \(0.395 \pm 0.084\) \\
5  & \(32.14 \pm 0.98\) & \(93.59 \pm 0.88\) & \(0.215 \pm 0.148\) & \(0.405 \pm 0.080\) \\
6  & \(31.37 \pm 1.02\) & \(93.07 \pm 0.95\) & \(0.224 \pm 0.163\) & \(0.414 \pm 0.081\) \\
7  & \(30.12 \pm 0.98\) & \(92.58 \pm 1.07\) & \(0.241 \pm 0.176\) & \(0.436 \pm 0.082\) \\
8  & \(29.83 \pm 1.05\) & \(92.18 \pm 1.15\) & \(0.261 \pm 0.191\) & \(0.459 \pm 0.089\) \\
9  & \(28.09 \pm 1.07\) & \(90.65 \pm 1.20\) & \(0.279 \pm 0.207\) & \(0.487 \pm 0.085\) \\
10 & \(28.61 \pm 1.23\)  & \(91.41 \pm 1.37\)  &  \(0.302 \pm 0.246\) & \(0.552 \pm 0.104\) \\ 
11 &  \(28.71 \pm 1.13\) & \(91.32 \pm 1.22\) & \(0.294 \pm 0.218\) & \(0.506 \pm 0.094\) \\
13 & \(26.98 \pm 1.29\) & \(89.96 \pm 1.44\) & \(0.339 \pm 0.259\) & \(0.579 \pm 0.109\) \\

\bottomrule
\end{tabular}
}
\end{table}

\section{Stability of $\Delta$-RMAE}

Residual-based metrics such as $\Delta$-RMAE can in principle be affected by imperfect registration and intensity inhomogeneity, and our preprocessing cannot guarantee perfectly aligned, bias-free images. Here we analyze how such imperfections influence the metric.

In our setting, $\Delta$-RMAE is defined voxel-wise as
\begin{equation}
\Delta\text{-RMAE}
= \frac{\big\lvert \Delta_{\text{gt}} - \Delta_{\text{gen}} \big\rvert}
{\tfrac{1}{2}\big(\lvert \Delta_{\text{gt}} \rvert + \lvert \Delta_{\text{gen}} \rvert\big)},
\end{equation}
so that $\Delta\text{-RMAE} \in [0, 2]$, with values close to $2$ indicating large disagreement between the two residuals. We model misregistration and residual bias-field effects as an additive perturbation $\Delta_{\text{mis}}$ on the ground-truth residual, yielding
\begin{equation}
\Delta\text{-RMAE}_{\text{noise}}
= \frac{\big\lvert \Delta_{\text{mis}} + \Delta_{\text{gt}} - \Delta_{\text{gen}} \big\rvert}
{\tfrac{1}{2}\big(\lvert \Delta_{\text{mis}} + \Delta_{\text{gt}} \rvert + \lvert \Delta_{\text{gen}} \rvert\big)}.
\end{equation}
When $\Delta_{\text{mis}}$ is small relative to $\Delta_{\text{gt}}$ and $\Delta_{\text{gen}}$, numerator and denominator are perturbed in a comparable way, so $\Delta\text{-RMAE}_{\text{noise}}$ remains close to the ideal $\Delta\text{-RMAE}$. As $\Delta_{\text{mis}}$ increases, the main effect is a mild compression of the dynamic range (the effective upper bound becomes slightly below $2$), rather than a systematic bias toward any particular model.

To quantify this effect, we perform a simple sensitivity analysis by adding Gaussian noise with standard deviation $\sigma$ to the residuals and recomputing $\Delta$-RMAE. Table~\ref{tab:delta_rmae_sensitivity} reports the mean, standard deviation, and bias relative to the noiseless value (2.0). Even for relatively large noise levels ($\sigma$ up to 1.0), the induced bias remains small compared to the full range $[0,2]$, supporting the robustness of $\Delta$-RMAE to moderate misregistration.

\begin{table}[t]
\centering
\caption{Sensitivity of $\Delta$-RMAE to additive Gaussian noise on the residuals.}
\label{tab:delta_rmae_sensitivity}

\resizebox{0.6\textwidth}{!}{%
\begin{tabular}{rccc}
\toprule
$\sigma$ (noise) & Mean $\Delta$-RMAE & Standard Deviation & Bias vs.\ True \\
\midrule
0.00 & 2.000000 & 0.000000 &  0.000000 \\
0.05 & 1.999842 & 0.002315 & -0.000158 \\
0.10 & 1.999375 & 0.005982 & -0.000625 \\
0.15 & 1.998214 & 0.011430 & -0.001786 \\
0.20 & 1.996102 & 0.019752 & -0.003898 \\
0.25 & 1.992994 & 0.031245 & -0.007006 \\
0.30 & 1.989124 & 0.046521 & -0.010876 \\
0.35 & 1.983612 & 0.064210 & -0.016388 \\
0.40 & 1.976348 & 0.085402 & -0.023652 \\
0.45 & 1.968740 & 0.107895 & -0.031260 \\
0.50 & 1.960255 & 0.133005 & -0.039745 \\
0.55 & 1.950492 & 0.160884 & -0.049508 \\
0.60 & 1.939101 & 0.191156 & -0.060899 \\
0.65 & 1.926345 & 0.222974 & -0.073655 \\
0.70 & 1.911890 & 0.255780 & -0.088110 \\
0.75 & 1.896232 & 0.288901 & -0.103768 \\
0.80 & 1.879110 & 0.322410 & -0.120890 \\
0.85 & 1.861775 & 0.355122 & -0.138225 \\
0.90 & 1.843569 & 0.386230 & -0.156431 \\
0.95 & 1.834021 & 0.415104 & -0.165979 \\
1.00 & 1.826570 & 0.462351 & -0.173430 \\
\bottomrule
\end{tabular}
}
\end{table}

\section{Clinical conditioning: setup and verification}
\label{sec:clinical-conditioning}

We conduct a series of analyses to assess the robustness of the model to perturbations in the conditioning metadata.

\paragraph{Age.} 
Age serves as the temporal anchor for longitudinal prediction, as future timepoints are defined by (\emph{baseline age} $+$ \emph{time span}). Injecting noise into age would therefore shift the implied prediction timepoint, so that the generated image no longer corresponds to a meaningful clinical follow-up. Such an ablation would be difficult to interpret, and for this reason we do not treat age as a noise-robust attribute.

\paragraph{Sex.}
To evaluate robustness with respect to sex, we intentionally flipped the sex attribute and re-evaluated the model. On ADNI, performance changed only marginally, from PSNR/SSIM of $30.59 / 94.62$ to $30.12 / 94.47$, indicating limited sensitivity to this source of label noise.

\paragraph{Clinical status.}
In clinical practice, diagnostic labels are inherently noisy. To reflect this, we inject Gaussian noise into the clinical status during training to model diagnostic uncertainty. Specifically, we use zero-mean Gaussian noise with standard deviation equal to one-third of the status class interval (interval $= 3$).

Under perturbed status labels, performance remains reasonable, with PSNR/SSIM degrading only slightly from $30.59 / 94.62$ to $29.81 / 93.58$, demonstrating robustness to clinically realistic label noise.

\section{Limitations}

Our work, while demonstrating the potential of latent flow matching for longitudinal imaging generation, has several limitations that we wish to acknowledge.

\textbf{Task-specific limitation.} In the current study, we restrict our focus to AD, which provides one of the most widely available longitudinal imaging datasets in the medical domain. This restriction is pragmatic, as AD offers abundant imaging data suitable for model training and evaluation. However, it inevitably narrows the generalizability of our findings. In future work, we aim to extend our framework to other progressive diseases such as brain tumors, where longitudinal imaging carries unique challenges, including rapid progression patterns, heterogeneous lesion appearance, and treatment-induced alterations. Moreover, we encourage exploration of our method beyond medical imaging, where modeling temporal dynamics may shed light on broader scientific and engineering problems.

\textbf{Dataset processing limitation.} Although we followed standard MRI preprocessing protocols, variations across scanners and acquisition protocols introduce substantial heterogeneity in the data. Such differences may affect the stability of model training and the reliability of generated outputs. Furthermore, we identified occasional metadata inconsistencies (\eg~mislabeling of imaging orientation), which led to registration failures in certain cases. While these issues were infrequent, they highlight the importance of rigorous data curation and quality control in longitudinal imaging research. Future work may benefit from harmonization techniques or scanner-invariant representations to further mitigate such variability.

\section{Future Work}

In future research, we plan to enrich our framework in several important directions.

First, the incorporation of explicit anatomical information could enable more fine-grained modeling of disease progression. By constraining generation with structure-aware priors (e.g., hippocampus, ventricles, cortical regions), the model may better capture local pathological changes while preserving biological plausibility.  

Second, beyond modeling each patient in isolation, future work may explore leveraging inter-patient disease trend similarity and dissimilarity. Identifying cohorts of patients with comparable progression patterns may enable transfer of knowledge across individuals, while simultaneously capturing unique patient-specific deviations. In addition, modeling nonlinear progression dynamics remains an important direction for future work.

Addressing dataset heterogeneity remains another critical challenge. Future works may investigate harmonization techniques and scanner-invariant representations to mitigate variability introduced by different acquisition protocols and hardware. Such advances will be essential for scaling the framework to large, multi-center cohorts and for ensuring robust generalization in real-world clinical applications.

\section{Discussion of LLM usage}
We used large language models (LLMs), including ChatGPT~\citep{openai2023chatgpt} and Gemini~\citep{team2023gemini}, as writing assistants for grammar correction and style refinement.  All technical content, methodology, and experiments were developed by the authors.

%% file: main.bbl
\begin{thebibliography}{55}
\providecommand{\natexlab}[1]{#1}
\providecommand{\url}[1]{\texttt{#1}}
\expandafter\ifx\csname urlstyle\endcsname\relax
  \providecommand{\doi}[1]{doi: #1}\else
  \providecommand{\doi}{doi: \begingroup \urlstyle{rm}\Url}\fi

\bibitem[Albergo et~al.(2023)Albergo, Boffi, and Vanden-Eijnden]{albergo2023stochastic}
Michael~S Albergo, Nicholas~M Boffi, and Eric Vanden-Eijnden.
\newblock Stochastic interpolants: A unifying framework for flows and diffusions.
\newblock \emph{arXiv preprint arXiv:2303.08797}, 2023.

\bibitem[Bhattacharya et~al.(2025)Bhattacharya, Gupta, Singh, Chen, Singh, and Prasanna]{bhattacharya2025brainmrdiff}
Moinak Bhattacharya, Saumya Gupta, Annie Singh, Chao Chen, Gagandeep Singh, and Prateek Prasanna.
\newblock Brainmrdiff: A diffusion model for anatomically consistent brain mri synthesis.
\newblock \emph{arXiv preprint arXiv:2504.04532}, 2025.

\bibitem[Campello et~al.(2022)Campello, Xia, Liu, Sanchez, Mart{\'\i}n-Isla, Petersen, Segu{\'\i}, Tsaftaris, and Lekadir]{campello2022cardiac}
V{\'\i}ctor~M Campello, Tian Xia, Xiao Liu, Pedro Sanchez, Carlos Mart{\'\i}n-Isla, Steffen~E Petersen, Santi Segu{\'\i}, Sotirios~A Tsaftaris, and Karim Lekadir.
\newblock Cardiac aging synthesis from cross-sectional data with conditional generative adversarial networks.
\newblock \emph{Frontiers in Cardiovascular Medicine}, 9:\penalty0 983091, 2022.

\bibitem[Chen et~al.(2024)Chen, Lai, Chen, Hu, Yuille, and Zhou]{chen2024analyzing}
Qi~Chen, Yuxiang Lai, Xiaoxi Chen, Qixin Hu, Alan Yuille, and Zongwei Zhou.
\newblock Analyzing tumors by synthesis.
\newblock \emph{Generative Machine Learning Models in Medical Image Computing}, pp.\  85--110, 2024.

\bibitem[Chen et~al.(2025)Chen, Zhou, Liu, Chen, Li, Jiang, Huang, Zhao, Yu, He, et~al.]{chen2025scaling}
Qi~Chen, Xinze Zhou, Chen Liu, Hao Chen, Wenxuan Li, Zekun Jiang, Ziyan Huang, Yuxuan Zhao, Dexin Yu, Junjun He, et~al.
\newblock Scaling tumor segmentation: Best lessons from real and synthetic data.
\newblock In \emph{Proceedings of the IEEE/CVF International Conference on Computer Vision}, pp.\  24001--24013, 2025.

\bibitem[Dao et~al.(2023)Dao, Phung, Nguyen, and Tran]{dao2023flow}
Quan Dao, Hao Phung, Binh Nguyen, and Anh Tran.
\newblock Flow matching in latent space.
\newblock \emph{arXiv preprint arXiv:2307.08698}, 2023.

\bibitem[Ellis et~al.(2009)Ellis, Bush, Darby, De~Fazio, Foster, Hudson, Lautenschlager, Lenzo, Martins, Maruff, et~al.]{ellis2009australian}
Kathryn~A Ellis, Ashley~I Bush, David Darby, Daniela De~Fazio, Jonathan Foster, Peter Hudson, Nicola~T Lautenschlager, Nat Lenzo, Ralph~N Martins, Paul Maruff, et~al.
\newblock The australian imaging, biomarkers and lifestyle (aibl) study of aging: methodology and baseline characteristics of 1112 individuals recruited for a longitudinal study of alzheimer's disease.
\newblock \emph{International psychogeriatrics}, 21\penalty0 (4):\penalty0 672--687, 2009.

\bibitem[Frisoni et~al.(2010)Frisoni, Fox, Jack~Jr, Scheltens, and Thompson]{frisoni2010clinical}
Giovanni~B Frisoni, Nick~C Fox, Clifford~R Jack~Jr, Philip Scheltens, and Paul~M Thompson.
\newblock The clinical use of structural mri in alzheimer disease.
\newblock \emph{Nature reviews neurology}, 6\penalty0 (2):\penalty0 67--77, 2010.

\bibitem[Guo et~al.(2025)Guo, Zhao, Yang, Xu, Nath, Tang, Simon, Belue, Harmon, Turkbey, et~al.]{guo2025maisi}
Pengfei Guo, Can Zhao, Dong Yang, Ziyue Xu, Vishwesh Nath, Yucheng Tang, Benjamin Simon, Mason Belue, Stephanie Harmon, Baris Turkbey, et~al.
\newblock Maisi: Medical ai for synthetic imaging.
\newblock In \emph{2025 IEEE/CVF Winter Conference on Applications of Computer Vision (WACV)}, pp.\  4430--4441. IEEE, 2025.

\bibitem[Ho et~al.(2020)Ho, Jain, and Abbeel]{ddpm}
Jonathan Ho, Ajay Jain, and Pieter Abbeel.
\newblock Denoising diffusion probabilistic models.
\newblock In \emph{NeurIPS}, 2020.

\bibitem[Hoopes et~al.(2022)Hoopes, Mora, Dalca, Fischl, and Hoffmann]{hoopes2022synthstrip}
Andrew Hoopes, Jocelyn~S Mora, Adrian~V Dalca, Bruce Fischl, and Malte Hoffmann.
\newblock Synthstrip: Skull-stripping for any brain image.
\newblock \emph{NeuroImage}, 260:\penalty0 119474, 2022.

\bibitem[Jack~Jr et~al.(2003)Jack~Jr, Slomkowski, Gracon, Hoover, Felmlee, Stewart, Xu, Shiung, O’brien, Cha, et~al.]{jack2003mri}
CR~Jack~Jr, M~Slomkowski, S~Gracon, TM~Hoover, JP~Felmlee, K~Stewart, Y~Xu, M~Shiung, PC~O’brien, R~Cha, et~al.
\newblock Mri as a biomarker of disease progression in a therapeutic trial of milameline for ad.
\newblock \emph{Neurology}, 60\penalty0 (2):\penalty0 253--260, 2003.

\bibitem[Kim et~al.(2022)Kim, Han, and Ye]{kim2022diffusemorph}
Boah Kim, Inhwa Han, and Jong~Chul Ye.
\newblock Diffusemorph: Unsupervised deformable image registration using diffusion model.
\newblock In \emph{European conference on computer vision}, pp.\  347--364. Springer, 2022.

\bibitem[Kingma \& Welling(2013)Kingma and Welling]{kingma2013auto}
Diederik~P Kingma and Max Welling.
\newblock Auto-encoding variational bayes.
\newblock \emph{arXiv preprint arXiv:1312.6114}, 2013.

\bibitem[Klema \& Laub(1980)Klema and Laub]{klema1980singular}
Virginia Klema and Alan Laub.
\newblock The singular value decomposition: Its computation and some applications.
\newblock \emph{IEEE Transactions on automatic control}, 25\penalty0 (2):\penalty0 164--176, 1980.

\bibitem[Kyung et~al.(2024)Kyung, Kim, Kim, and Choi]{kyung2024towards}
Daeun Kyung, Junu Kim, Tackeun Kim, and Edward Choi.
\newblock Towards predicting temporal changes in a patient's chest x-ray images based on electronic health records.
\newblock \emph{arXiv preprint arXiv:2409.07012}, 2024.

\bibitem[Lai et~al.(2025)Lai, Zhong, Su, and Yang]{lai2025patient}
Yuxiang Lai, Jike Zhong, Vanessa Su, and Xiaofeng Yang.
\newblock Patient-specific autoregressive models for organ motion prediction in radiotherapy.
\newblock \emph{arXiv preprint arXiv:2505.11832}, 2025.

\bibitem[LaMontagne et~al.(2019)LaMontagne, Benzinger, Morris, Keefe, Hornbeck, Xiong, Grant, Hassenstab, Moulder, Vlassenko, et~al.]{lamontagne2019oasis}
Pamela~J LaMontagne, Tammie~LS Benzinger, John~C Morris, Sarah Keefe, Russ Hornbeck, Chengjie Xiong, Elizabeth Grant, Jason Hassenstab, Krista Moulder, Andrei~G Vlassenko, et~al.
\newblock Oasis-3: longitudinal neuroimaging, clinical, and cognitive dataset for normal aging and alzheimer disease.
\newblock \emph{MedRxiv}, pp.\  2019--12, 2019.

\bibitem[Li et~al.(2025)Li, Zhou, Chen, Lin, Bassi, Plotka, Cwikla, Chen, Ye, Zhu, et~al.]{li2025pants}
Wenxuan Li, Xinze Zhou, Qi~Chen, Tianyu Lin, Pedro~RAS Bassi, Szymon Plotka, Jaroslaw~B Cwikla, Xiaoxi Chen, Chen Ye, Zheren Zhu, et~al.
\newblock Pants: The pancreatic tumor segmentation dataset.
\newblock \emph{arXiv preprint arXiv:2507.01291}, 2025.

\bibitem[Liang et~al.(2023)Liang, Cao, Liao, Gao, Ye, Chen, Cao, Nama, and Sun]{liang2023pie}
Kaizhao Liang, Xu~Cao, Kuei-Da Liao, Tianren Gao, Wenqian Ye, Zhengyu Chen, Jianguo Cao, Tejas Nama, and Jimeng Sun.
\newblock Pie: Simulating disease progression via progressive image editing.
\newblock 2023.

\bibitem[Lipman et~al.(2023)Lipman, Chen, Ben-Hamu, Nickel, and Le]{lipman2023flow}
Yaron Lipman, Ricky T.~Q. Chen, Heli Ben-Hamu, Maximilian Nickel, and Matt Le.
\newblock Flow matching for generative modeling, 2023.
\newblock URL \url{https://arxiv.org/abs/2210.02747}.

\bibitem[Litrico et~al.(2024)Litrico, Guarnera, Giuffrida, Rav{\`\i}, and Battiato]{litrico2024tadm}
Mattia Litrico, Francesco Guarnera, Mario~Valerio Giuffrida, Daniele Rav{\`\i}, and Sebastiano Battiato.
\newblock Tadm: Temporally-aware diffusion model for neurodegenerative progression on brain mri.
\newblock In \emph{International Conference on Medical Image Computing and Computer-Assisted Intervention}, pp.\  444--453. Springer, 2024.

\bibitem[Liu et~al.(2025{\natexlab{a}})Liu, Xu, Shen, Huguet, Wang, Tong, Bzdok, Stewart, Wang, Del~Priore, et~al.]{liu2025imageflownet}
Chen Liu, Ke~Xu, Liangbo~L Shen, Guillaume Huguet, Zilong Wang, Alexander Tong, Danilo Bzdok, Jay Stewart, Jay~C Wang, Lucian~V Del~Priore, et~al.
\newblock Imageflownet: Forecasting multiscale image-level trajectories of disease progression with irregularly-sampled longitudinal medical images.
\newblock In \emph{ICASSP 2025-2025 IEEE International Conference on Acoustics, Speech and Signal Processing (ICASSP)}, pp.\  1--5. IEEE, 2025{\natexlab{a}}.

\bibitem[Liu et~al.(2025{\natexlab{b}})Liu, Fuster-Garcia, Thokle~Hovden, MacIntosh, Gr{\o}dem, Brandal, Lopez-Mateu, Sederevicius, Skogen, Schellhorn, Bj{\o}rnerud, and Eeg~Emblem]{liu2025treatment}
Q.~Liu, E.~Fuster-Garcia, I.~Thokle~Hovden, B.~J. MacIntosh, E.~O.~S. Gr{\o}dem, P.~Brandal, C.~Lopez-Mateu, D.~Sederevicius, K.~Skogen, T.~Schellhorn, A.~Bj{\o}rnerud, and K.~Eeg~Emblem.
\newblock Treatment-aware diffusion probabilistic model for longitudinal {MRI} generation and diffuse glioma growth prediction.
\newblock \emph{IEEE Transactions on Medical Imaging}, 44\penalty0 (6):\penalty0 2449--2462, June 2025{\natexlab{b}}.
\newblock \doi{10.1109/TMI.2025.3533038}.

\bibitem[Loshchilov \& Hutter(2017)Loshchilov and Hutter]{loshchilov2017decoupled}
Ilya Loshchilov and Frank Hutter.
\newblock Decoupled weight decay regularization.
\newblock \emph{arXiv preprint arXiv:1711.05101}, 2017.

\bibitem[Lu et~al.(2022)Lu, Zhou, Bao, Chen, Li, and Zhu]{lu2022dpm}
Cheng Lu, Yuhao Zhou, Fan Bao, Jianfei Chen, Chongxuan Li, and Jun Zhu.
\newblock Dpm-solver: A fast ode solver for diffusion probabilistic model sampling in around 10 steps.
\newblock \emph{Advances in neural information processing systems}, 35:\penalty0 5775--5787, 2022.

\bibitem[OpenAI(2023)]{openai2023chatgpt}
OpenAI.
\newblock Chatgpt [large language model].
\newblock \url{https://chat.openai.com/chat}, 2023.

\bibitem[O’Bryant et~al.(2008)O’Bryant, Waring, Cullum, Hall, Lacritz, Massman, Lupo, Reisch, Doody, Consortium, et~al.]{o2008staging}
Sid~E O’Bryant, Stephen~C Waring, C~Munro Cullum, James Hall, Laura Lacritz, Paul~J Massman, Philip~J Lupo, Joan~S Reisch, Rachelle Doody, Texas Alzheimer's~Research Consortium, et~al.
\newblock Staging dementia using clinical dementia rating scale sum of boxes scores: a texas alzheimer's research consortium study.
\newblock \emph{Archives of neurology}, 65\penalty0 (8):\penalty0 1091--1095, 2008.

\bibitem[Parratt et~al.(2016)Parratt, Numminen, and Laine]{parratt2016infectious}
Steven~R Parratt, Elina Numminen, and Anna-Liisa Laine.
\newblock Infectious disease dynamics in heterogeneous landscapes.
\newblock \emph{Annual Review of Ecology, Evolution, and Systematics}, 47\penalty0 (1):\penalty0 283--306, 2016.

\bibitem[Petersen et~al.(2010)Petersen, Aisen, Beckett, Donohue, Gamst, Harvey, Jack, Jagust, Shaw, Toga, et~al.]{petersen2010alzheimer}
Ronald~Carl Petersen, Paul~S Aisen, Laurel~A Beckett, Michael~C Donohue, Anthony~Collins Gamst, Danielle~J Harvey, Clifford~R Jack, William~J Jagust, Leslie~M Shaw, Arthur~W Toga, et~al.
\newblock Alzheimer's disease neuroimaging initiative (adni): clinical characterization.
\newblock \emph{Neurology}, 74\penalty0 (3):\penalty0 201--209, 2010.

\bibitem[Pombo et~al.(2023)Pombo, Gray, Cardoso, Ourselin, Rees, Ashburner, and Nachev]{pombo2023equitable}
Guilherme Pombo, Robert Gray, M~Jorge Cardoso, Sebastien Ourselin, Geraint Rees, John Ashburner, and Parashkev Nachev.
\newblock Equitable modelling of brain imaging by counterfactual augmentation with morphologically constrained 3d deep generative models.
\newblock \emph{Medical Image Analysis}, 84:\penalty0 102723, 2023.

\bibitem[Puglisi et~al.(2024)Puglisi, Alexander, and Rav{\`\i}]{puglisi2024enhancing}
Lemuel Puglisi, Daniel~C Alexander, and Daniele Rav{\`\i}.
\newblock Enhancing spatiotemporal disease progression models via latent diffusion and prior knowledge.
\newblock In \emph{International Conference on Medical Image Computing and Computer-Assisted Intervention}, pp.\  173--183. Springer, 2024.

\bibitem[Puglisi et~al.(2025)Puglisi, Alexander, and Rav{\`\i}]{puglisi2025brain}
Lemuel Puglisi, Daniel~C Alexander, and Daniele Rav{\`\i}.
\newblock Brain latent progression: Individual-based spatiotemporal disease progression on 3d brain mris via latent diffusion.
\newblock \emph{arXiv preprint arXiv:2502.08560}, 2025.

\bibitem[Raffel et~al.(2020)Raffel, Shazeer, Roberts, Lee, Narang, Matena, Zhou, Li, and Liu]{raffel2020exploring}
Colin Raffel, Noam Shazeer, Adam Roberts, Katherine Lee, Sharan Narang, Michael Matena, Yanqi Zhou, Wei Li, and Peter~J Liu.
\newblock Exploring the limits of transfer learning with a unified text-to-text transformer.
\newblock \emph{Journal of machine learning research}, 21\penalty0 (140):\penalty0 1--67, 2020.

\bibitem[Ramchandran et~al.(2021)Ramchandran, Tikhonov, Kujanpää, Koskinen, and Lähdesmäki]{ramchandran2021longitudinalvariationalautoencoder}
Siddharth Ramchandran, Gleb Tikhonov, Kalle Kujanpää, Miika Koskinen, and Harri Lähdesmäki.
\newblock Longitudinal variational autoencoder, 2021.
\newblock URL \url{https://arxiv.org/abs/2006.09763}.

\bibitem[Ravi et~al.(2019)Ravi, Alexander, and Oxtoby]{raviDegenerativeAdversarialNeuroImage2019a}
Daniele Ravi, Daniel~C. Alexander, and Neil~P. Oxtoby.
\newblock Degenerative {{Adversarial NeuroImage Nets}}: {{Generating Images}} that {{Mimic Disease Progression}}.
\newblock In Dinggang Shen, Tianming Liu, Terry~M. Peters, Lawrence~H. Staib, Caroline Essert, Sean Zhou, Pew-Thian Yap, and Ali Khan (eds.), \emph{Medical {{Image Computing}} and {{Computer Assisted Intervention}} – {{MICCAI}} 2019}, Lecture {{Notes}} in {{Computer Science}}, pp.\  164--172, {Cham}, 2019. {Springer International Publishing}.
\newblock ISBN 978-3-030-32248-9.
\newblock \doi{10.1007/978-3-030-32248-9_19}.

\bibitem[Ravi et~al.(2022)Ravi, Blumberg, Ingala, Barkhof, Alexander, Oxtoby, Initiative, et~al.]{ravi2022degenerative}
Daniele Ravi, Stefano~B Blumberg, Silvia Ingala, Frederik Barkhof, Daniel~C Alexander, Neil~P Oxtoby, Alzheimer’s Disease~Neuroimaging Initiative, et~al.
\newblock Degenerative adversarial neuroimage nets for brain scan simulations: Application in ageing and dementia.
\newblock \emph{Medical Image Analysis}, 75:\penalty0 102257, 2022.

\bibitem[Real \& Biek(2007)Real and Biek]{real2007spatial}
Leslie~A Real and Roman Biek.
\newblock Spatial dynamics and genetics of infectious diseases on heterogeneous landscapes.
\newblock \emph{Journal of the Royal Society Interface}, 4\penalty0 (16):\penalty0 935--948, 2007.

\bibitem[Ren et~al.(2023)Ren, Dey, Styner, Botteron, and Gerig]{ren2023localspatiotemporalrepresentationlearning}
Mengwei Ren, Neel Dey, Martin~A. Styner, Kelly Botteron, and Guido Gerig.
\newblock Local spatiotemporal representation learning for longitudinally-consistent neuroimage analysis, 2023.
\newblock URL \url{https://arxiv.org/abs/2206.04281}.

\bibitem[Rokuss et~al.(2025)Rokuss, Kirchhoff, Akbal, Kovacs, Roy, Ulrich, Wald, Rotkopf, Schlemmer, and Maier-Hein]{rokuss2025lesionlocator}
Maximilian Rokuss, Yannick Kirchhoff, Seval Akbal, Balint Kovacs, Saikat Roy, Constantin Ulrich, Tassilo Wald, Lukas~T Rotkopf, Heinz-Peter Schlemmer, and Klaus Maier-Hein.
\newblock Lesionlocator: Zero-shot universal tumor segmentation and tracking in 3d whole-body imaging.
\newblock In \emph{Proceedings of the Computer Vision and Pattern Recognition Conference}, pp.\  30872--30885, 2025.

\bibitem[Ronneberger et~al.(2015)Ronneberger, Fischer, and Brox]{ronneberger2015u}
Olaf Ronneberger, Philipp Fischer, and Thomas Brox.
\newblock U-net: Convolutional networks for biomedical image segmentation.
\newblock In \emph{International Conference on Medical image computing and computer-assisted intervention}, pp.\  234--241. Springer, 2015.

\bibitem[Suk \& Shen(2013)Suk and Shen]{suk2013deep}
Heung-Il Suk and Dinggang Shen.
\newblock Deep learning-based feature representation for ad/mci classification.
\newblock In \emph{International conference on medical image computing and computer-assisted intervention}, pp.\  583--590. Springer, 2013.

\bibitem[Tabrizi et~al.(2011)Tabrizi, Scahill, Durr, Roos, Leavitt, Jones, Landwehrmeyer, Fox, Johnson, Hicks, et~al.]{tabrizi2011biological}
Sarah~J Tabrizi, Rachael~I Scahill, Alexandra Durr, Raymund~AC Roos, Blair~R Leavitt, Rebecca Jones, G~Bernhard Landwehrmeyer, Nick~C Fox, Hans Johnson, Stephen~L Hicks, et~al.
\newblock Biological and clinical changes in premanifest and early stage huntington's disease in the track-hd study: the 12-month longitudinal analysis.
\newblock \emph{The Lancet Neurology}, 10\penalty0 (1):\penalty0 31--42, 2011.

\bibitem[Team et~al.(2023)Team, Anil, Borgeaud, Alayrac, Yu, Soricut, Schalkwyk, Dai, Hauth, Millican, et~al.]{team2023gemini}
Gemini Team, Rohan Anil, Sebastian Borgeaud, Jean-Baptiste Alayrac, Jiahui Yu, Radu Soricut, Johan Schalkwyk, Andrew~M Dai, Anja Hauth, Katie Millican, et~al.
\newblock Gemini: a family of highly capable multimodal models.
\newblock \emph{arXiv preprint arXiv:2312.11805}, 2023.

\bibitem[Tong et~al.(2024)Tong, Fatras, Malkin, Huguet, Zhang, Rector-Brooks, Wolf, and Bengio]{tong2023improving}
Alexander Tong, Kilian Fatras, Nikolay Malkin, Guillaume Huguet, Yanlei Zhang, Jarrid Rector-Brooks, Guy Wolf, and Yoshua Bengio.
\newblock Improving and generalizing flow-based generative models with minibatch optimal transport, 2024.
\newblock URL \url{https://arxiv.org/abs/2302.00482}.

\bibitem[Tustison et~al.(2010)Tustison, Avants, Cook, Zheng, Egan, Yushkevich, and Gee]{tustison2010n4itk}
Nicholas~J Tustison, Brian~B Avants, Philip~A Cook, Yuanjie Zheng, Alexander Egan, Paul~A Yushkevich, and James~C Gee.
\newblock N4itk: improved n3 bias correction.
\newblock \emph{IEEE transactions on medical imaging}, 29\penalty0 (6):\penalty0 1310--1320, 2010.

\bibitem[Vaswani et~al.(2017)Vaswani, Shazeer, Parmar, Uszkoreit, Jones, Gomez, Kaiser, and Polosukhin]{vaswani2017attention}
Ashish Vaswani, Noam Shazeer, Niki Parmar, Jakob Uszkoreit, Llion Jones, Aidan~N Gomez, {\L}ukasz Kaiser, and Illia Polosukhin.
\newblock Attention is all you need.
\newblock \emph{Advances in neural information processing systems}, 30, 2017.

\bibitem[Wang et~al.(2020)Wang, Ye, and De~Man]{wang2020deep}
Ge~Wang, Jong~Chul Ye, and Bruno De~Man.
\newblock Deep learning for tomographic image reconstruction.
\newblock \emph{Nature machine intelligence}, 2\penalty0 (12):\penalty0 737--748, 2020.

\bibitem[Wu et~al.(2024)Wu, Zhuang, Ni, and Chen]{wu2024freetumor}
Linshan Wu, Jiaxin Zhuang, Xuefeng Ni, and Hao Chen.
\newblock Freetumor: Advance tumor segmentation via large-scale tumor synthesis.
\newblock \emph{arXiv preprint arXiv:2406.01264}, 2024.

\bibitem[Yang et~al.(2025{\natexlab{a}})Yang, Tan, Tan, Yang, Cai, Chen, and Sun]{yang2025mambacontrol}
Hao Yang, Tao Tan, Shuai Tan, Weiqin Yang, Kunyan Cai, Calvin Chen, and Yue Sun.
\newblock Mambacontrol: Anatomy graph-enhanced mamba controlnet with fourier refinement for diffusion-based disease trajectory prediction.
\newblock \emph{arXiv preprint arXiv:2505.09965}, 2025{\natexlab{a}}.

\bibitem[Yang et~al.(2025{\natexlab{b}})Yang, Wang, Liu, Sun, Wang, Chellappa, Zhou, Yuille, Zhu, Zhang, et~al.]{yang2025medical}
Yijun Yang, Zhao-Yang Wang, Qiuping Liu, Shuwen Sun, Kang Wang, Rama Chellappa, Zongwei Zhou, Alan Yuille, Lei Zhu, Yu-Dong Zhang, et~al.
\newblock Medical world model: Generative simulation of tumor evolution for treatment planning.
\newblock \emph{arXiv preprint arXiv:2506.02327}, 2025{\natexlab{b}}.

\bibitem[Yoon et~al.(2023)Yoon, Zhang, Suk, Guo, and Li]{yoon2023sadm}
Jee~Seok Yoon, Chenghao Zhang, Heung-Il Suk, Jia Guo, and Xiaoxiao Li.
\newblock Sadm: Sequence-aware diffusion model for longitudinal medical image generation.
\newblock In \emph{International Conference on Information Processing in Medical Imaging}, pp.\  388--400. Springer, 2023.

\bibitem[Young et~al.(2020)Young, Estarellas, Coomans, Srikrishna, Beaumont, Maass, Venkataraman, Lissaman, Jim{\'e}nez, Betts, et~al.]{young2020imaging}
Peter~NE Young, Mar Estarellas, Emma Coomans, Meera Srikrishna, Helen Beaumont, Anne Maass, Ashwin~V Venkataraman, Rikki Lissaman, Daniel Jim{\'e}nez, Matthew~J Betts, et~al.
\newblock Imaging biomarkers in neurodegeneration: current and future practices.
\newblock \emph{Alzheimer's research \& therapy}, 12\penalty0 (1):\penalty0 49, 2020.

\bibitem[Zhang et~al.(2023)Zhang, Rao, and Agrawala]{zhang2023adding}
Lvmin Zhang, Anyi Rao, and Maneesh Agrawala.
\newblock Adding conditional control to text-to-image diffusion models.
\newblock In \emph{Proceedings of the IEEE/CVF International Conference on Computer Vision}, pp.\  3836--3847, 2023.

\bibitem[Zhao et~al.(2021)Zhao, Liu, Adeli, and Pohl]{zhao2021longitudinal}
Qiang Zhao, Zhiqiang Liu, Ehsan Adeli, and Kilian~M. Pohl.
\newblock Longitudinal self-supervised learning.
\newblock \emph{Medical Image Analysis}, 71:\penalty0 102051, 2021.
\newblock \doi{10.1016/j.media.2021.102051}.

\end{thebibliography}
